%% file: arxiv.tex
\documentclass{article} % For LaTeX2e
\usepackage{iclr2027_conference,times}

\input{math_commands.tex}

\usepackage[accsupp]{axessibility}

\usepackage{graphicx}
\usepackage{booktabs}
\usepackage[table]{xcolor}
\usepackage{multirow}

\usepackage{hyperref}
\usepackage{cleveref}
\usepackage{url}

\crefname{figure}{Fig. }{Figs. }
\crefname{table}{Tab. }{Tabs. }
\crefname{section}{Sec. }{Secs. }
\Crefname{Section}{Section }{Sections }

\usepackage{orcidlink}
\usepackage{pifont}
\usepackage{lmodern}

\definecolor{dark_skyblue}{RGB}{102, 138, 209}
\definecolor{skyblue}{RGB}{204,215,237}
\definecolor{lightblue}{RGB}{230,235,245}
\definecolor{new_lightblue}{RGB}{156,175,214}
\definecolor{lightgray}{RGB}{240,240,240}
\definecolor{crimson}{RGB}{220,20,60}

\title{SpatialBlock: Enhancing Spatial Intelligence in LVLMs via Synthetic Block-Stacking Problem}

\author{Soohyun Ryu$^*$ \& Sohee Kim$^*$ \\
KAIST \\
South Korea \\
\texttt{\char`\{rsoohyun, joyhee\char`\}@kaist.ac.kr} \\
\And
Eunho Yang$^\dagger$ \\
KAIST \& AITRICS \\
South Korea \\
\texttt{yangeh@kaist.ac.kr}
}

\iclrfinalcopy % Uncomment for camera-ready version, but NOT for submission.
\begin{document}

\maketitle

\begingroup
\renewcommand{\thefootnote}{}
\footnotetext[0]{\textsuperscript{$*$} Equal contribution, \textsuperscript{$\dagger$} Corresponding Author.}
\endgroup

\begin{abstract}
  Large Vision-Language Models (LVLMs) have achieved strong performance on diverse visual tasks, yet their ability to reconstruct and reason about the 3D structure of the scene depicted in 2D images -- referred to as \textit{spatial intelligence} -- remains limited. Existing approaches attempt to address this gap by using real-scene spatial question answering datasets that require dense geometric annotations. However, constructing such labels is costly, time-consuming, and often noisy due to reliance on external perception modules. In this work, we propose a novel paradigm inspired by human cognitive development: learning foundational spatial skills through structured block-manipulation tasks. We introduce \textbf{SpatialBlock-15k}, a synthetic dataset of 15,000 block-stacking problems covering 3D-to-2D projection, viewpoint transformation, and structural combination. The dataset further incorporates controlled color modulation as visual cues to encourage anchor-based reasoning in visually complex conditions. Experiments demonstrate that LVLMs trained on our dataset through either direct answering or reasoning-based prediction significantly outperform baselines and generalize to real-world spatial tasks, despite the dataset’s synthetic and compact nature. Code and data are available at \url{https://github.com/rsoohyun/SpatialBlock}.
\end{abstract}
% \vspace{-0.6cm}

\section{Introduction}
\label{sec:intro}
Large Vision–Language Models (LVLMs)~\citep{wang2024qwen2,bai2025qwen3,zhu2025internvl3} have demonstrated remarkable performance across a wide range of 2D image understanding and reasoning tasks. Despite this progress, their ability to mentally reconstruct the 3D structure of a scene from 2D images -- namely, \textit{spatial intelligence} -- remains limited. This shortcoming poses a fundamental bottleneck for deploying LVLMs in real-world applications that require robust spatial reasoning, such as autonomous driving~\citep{tian2025nuscenes} and robotics~\citep{feng2025seeing}.

To address this limitation, prior work has extended model architectures by adding a specialized module for learning 3D features~\citep{cheng2024spatialrgpt,wu2025spatial} or designed tasks that directly require spatial reasoning~\citep{yang2025thinking,yin2025spatial} to encourage models to learn such capabilities. However, these approaches typically rely on real-world scene question answering datasets that require dense geometric and object-level annotations of the visual context. Constructing such dense labels is not only time-consuming and costly, but also noisy due to the frequent reliance on external modules (\emph{e.g.}, segmentation or depth estimation models), thereby limiting their reliability and scalability. 

\begin{figure}[!t]
    \centering
    \includegraphics[width=\linewidth]{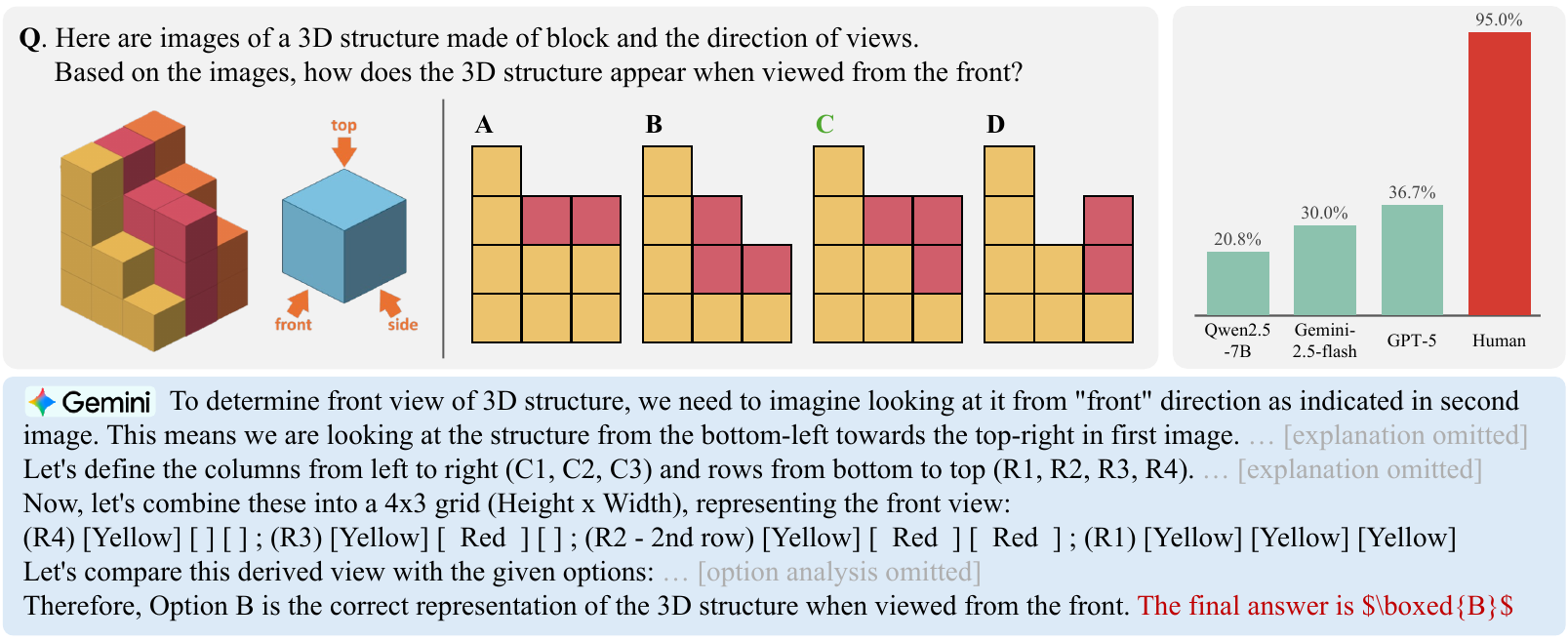}
    % \vspace{-0.6cm}
    \caption{\textbf{Performance Gap Between Humans and LVLMs in Block-Stacking Problem}. The top row shows a block-stacking task that requires predicting a 2D projection from a given 3D structure. While humans achieve near-perfect accuracy, state-of-the-art LVLMs struggle even with basic spatial reasoning problems. Leading proprietary models exhibit flawed spatial logic, often producing incorrect interpretations of the structure.}
    \label{fig:intro}
    % \vspace{-0.5cm}
\end{figure}

In contrast to methods that depend on annotation-heavy real-scene data, we pursue a more efficient alternative inspired by the way humans develop spatial intelligence. Developmental psychology~\citep{levine2012early,jirout2015building} suggests that foundational spatial abilities are often cultivated through structured manipulative tools. Specifically, block play has been widely shown to enhance essential components of spatial cognition, including spatial integration and mental simulation~\citep{caldera1999children, casey2008development}. Despite its simplicity, state-of-the-art LVLMs struggle with such problems, suggesting that they lack even fundamental levels of spatial intelligence (\cref{fig:intro}).

Motivated by this observation, we propose \textbf{SpatialBlock-15k}, a novel synthetic dataset of block-stacking problems designed to improve spatial intelligence. The dataset comprises three categories of spatial reasoning tasks: (1) 3D-to-2D projection, (2) viewpoint transformation, and (3) structural combination. To better approximate real-world spatial reasoning scenarios, where humans interpret scenes by anchoring on a specific object and reasoning relationships relative to it, we introduce color modulation as a visual cue, encouraging models to identify task-relevant elements in visually complex scenes. Unlike costly real-scene datasets, our dataset can be generated synthetically, enabling clean, scalable, and cost-effective data construction.

Experimental results show that LVLMs trained on SpatialBlock-15k, with either direct answer prediction or reasoning-based prediction, significantly outperform baseline models, even though the training data are entirely synthetic and relatively small in scale. These findings suggest that targeted training on fundamental spatial reasoning tasks can effectively enhance the spatial intelligence of LVLMs and provide a scalable alternative to annotation-heavy real-scene supervision.

In summary, our contribution is three-fold:
% \vspace{-0.2cm}
\begin{itemize}
    \item We introduce a new perspective on enhancing spatial intelligence in LVLMs by shifting from annotation-heavy real-scene supervision to foundational spatial skill learning through structured synthetic tasks.
    \item We present \textbf{SpatialBlock-15k}, a scalable synthetic dataset of block-stacking problems spanning 3D-to-2D projection, viewpoint transformation, and structural combination, enhanced with controlled color cues for task-relevant reasoning.
    \item We show that training LVLMs on SpatialBlock-15k significantly improves spatial reasoning performance and generalizes effectively to real-world scenes, despite relying solely on synthetic data.
\end{itemize}

\section{Related Work}
\label{sec:related_work}

\subsection{Visual Spatial Reasoning in LVLMs}
% \vspace{-0.2cm}
Recent efforts to improve spatial intelligence in LVLMs focus on building a dedicated encoder or proposing a specialized training strategy. On the architectural side, some approaches modify the model by introducing spatial tokens into the vision encoder~\citep{tong2024cambrian, lou2025llava}, incorporating depth-aware plugin modules~\citep{cheng2024spatialrgpt}, or adding spatial encoders to learn 3D features~\citep{wu2025spatial}. Other research focuses on the training process, designing hierarchical schemes that transition from spatial perception to complex reasoning~\citep{ma2025spatialreasoner, li2026spatialladder, liu2026spatial} or encouraging the model to output cognitive maps that encode object position and orientation~\citep{yang2025thinking, yin2025spatial}. Despite their effectiveness, these approaches generally require real-scene annotations or pseudo-3D signals from external modules, which limits their scalability and robustness. 

% \vspace{-0.2cm}
\subsection{Training Datasets for Spatial Intelligence}
% \vspace{-0.2cm}
Various training datasets have been proposed to enhance the spatial intelligence of LVLMs, ranging from simple spatial relations to complex multi-step reasoning. Early benchmarks~\citep{chen2024spatialvlm, ma2025spatialllm} leverage external models (\emph{e.g.}, object detection, segmentation, depth, or pose estimation) to extract 3D information and generate low-level QA pairs such as distance, orientation, and spatial relations. To foster higher-level logic, recent works~\citep{yang2025thinking,ouyang2025spacer} use 3D annotated video data to construct relative relation and direction reasoning, or more sophisticated tasks such as route planning or spatio-temporal appearance ordering through human annotation. MindCube~\citep{yin2025spatial} focuses on spatial mental modeling, which requires synthesizing a full scene from partial views or inferring arrangements under perspective shifts. Despite these advancements, a significant bottleneck remains: these approaches rely on dense, real-scene 3D annotations, which are labor-intensive and inherently noisy, necessitating costly human annotation. 

\section{SpatialBlock-15k}
\label{sec:dataset}

We introduce \textbf{SpatialBlock-15k}, a scalable synthetic dataset designed to systematically probe and enhance foundational spatial abilities, moving away from the conventional annotation-intensive paradigm that relies on dense real-scene 3D labels. Specifically, we utilize block-stacking problems, which play a pivotal role in the early stages of human spatial cognitive development (\cref{subsec:dataset_noncolor}). Furthermore, we incorporate controlled color variations as visual cues to better reflect real-world spatial reasoning conditions (\cref{subsec:dataset_color}).

\begin{figure}[!t]
    \centering
    \includegraphics[width=\linewidth]{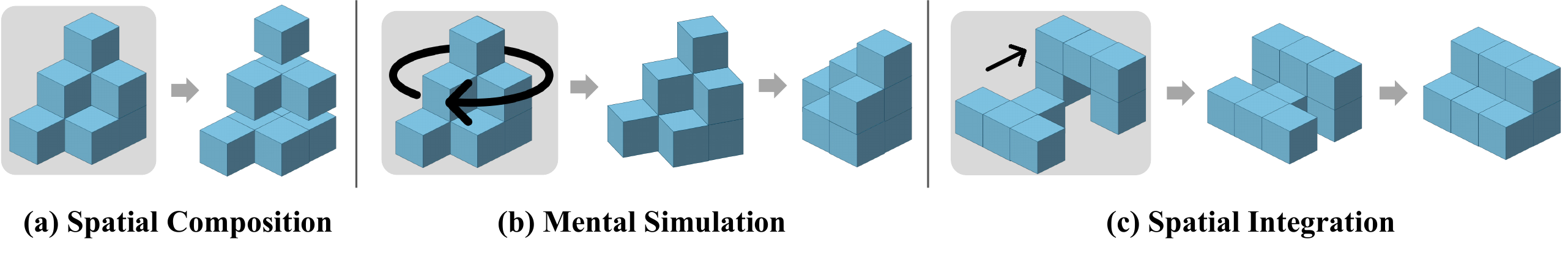}
    % \vspace{-0.8cm}
    \caption{\textbf{Spatial Abilities Underlying Block-Stacking Reasoning}. Solving block-stacking tasks involves (a) inferring latent 3D composition, (b) reasoning under transformation, and (c) integrating multiple components into a coherent structure.}
    \label{fig:block2SI}
    % \vspace{-0.5cm}
\end{figure}

\subsection{Block-Stacking Problems for Spatial Intelligence} 
\label{subsec:dataset_noncolor}

Activities involving block-stacking require structural spatial understanding that extends beyond superficial visual pattern recognition. As illustrated in \cref{fig:block2SI}, solving such problems requires (a) spatial composition, the ability to reconstruct the complete 3D structure by deconstructing the visible configuration and inferring occluded blocks that must exist to support it; (b) mental simulation, the capacity to simulate the outcome of transformations applied to a 3D structure; and (c) spatial integration, the ability to anticipate the resulting structure when multiple components are combined. Together, these abilities reflect core dimensions of spatial intelligence. 

\begin{figure}[!t]
    \centering
    \includegraphics[width=\linewidth]{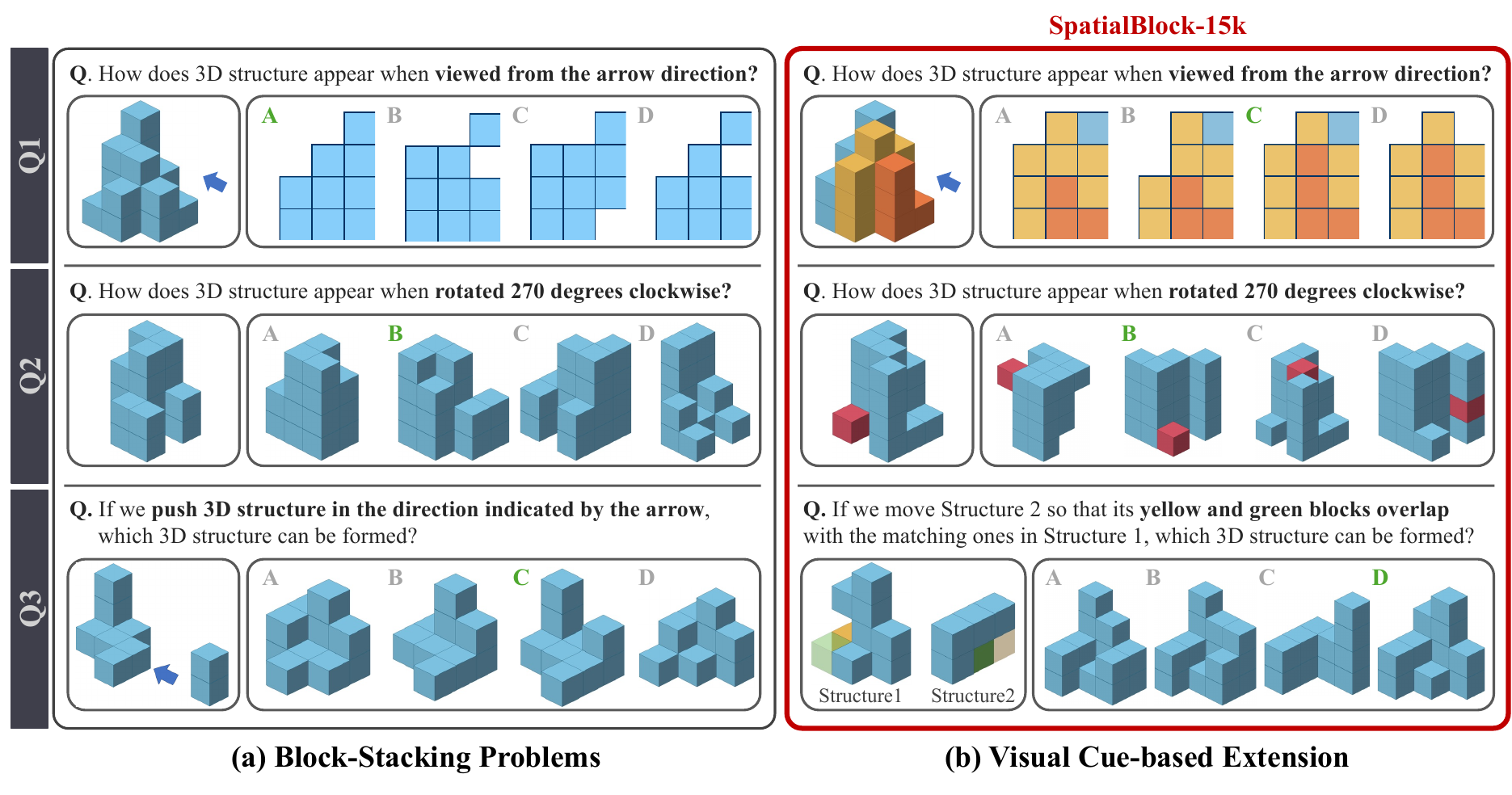}
    % \vspace{-0.8cm}
    \caption{\textbf{SpatialBlock-15k}. (a) Three block-stacking tasks: 3D-to-2D projection (Q1), viewpoint transformation (Q2), and structural combination (Q3). (b) Visual cue-based extension with controlled color variations, where colors encode depth ordering (Q1), preserved structural roles across transformations (Q2), and correspondence cues for integration (Q3).}
    \label{fig:spatialblock}
    % \vspace{-0.3cm}
\end{figure}

Building on these three abilities, we define three question types, each targeting a distinct aspect of spatial reasoning (\cref{fig:spatialblock}(a)): 

\begin{itemize}
    \item[\textbf{Q1:}] \textbf{3D-to-2D projection} 
    This task requires predicting the 2D appearance of a 3D structure from a specific viewing direction. Solving it demands the ability to internally reconstruct the 3D configuration and mentally transform it to the target viewpoint. The model must additionally perform depth-aware occlusion reasoning to determine which blocks remain visible in the final 2D projection.

    \item[\textbf{Q2:}] \textbf{Viewpoint transformation} 
    The problem inquires how a given 3D structure appears under self-rotation or viewpoint changes. The model should preserve structural consistency across transformations, maintaining correct relative positions of all components without distortion, disappearance, or unintended displacement.
    
    \item[\textbf{Q3:}] \textbf{Structural combination} 
    This question asks how the resultant integration is formed when two 3D structures are combined. It requires simulating how individual components interact, identifying contact interfaces, and subsequently inferring the coherent global structure that emerges from their integration. 
\end{itemize}

Collectively, these question types form the foundation of our SpatialBlock-15k dataset, providing a structured framework for evaluating and training spatial reasoning in LVLMs.

\subsection{Visual Cue-based Extension}
\label{subsec:dataset_color}

In the previous section, we constructed tasks that require understanding 3D structures composed of single-color blocks. Since color does not provide a discriminative feature, these tasks demand that the model focus on purely geometric relationships, such as the relative positions of blocks and appearance changes under viewpoint transformations.

However, real-world scenes typically consist of complex environments with diverse objects, rather than visually uniform ones. Given this intricacy, humans tend to analyze the global scene configurations by selecting a specific object as an anchor and interpreting relationships relative to it~\citep{land2001ways}. That is, visually salient cues serve as the reference points to infer changes and relations. 

Motivated by this cognitive characteristic, we extend the monochromatic block-based tasks introduced in \cref{subsec:dataset_noncolor} by incorporating color as an additional cue. Here, color is not used merely for visual diversity but as a functional guidance that reveals depth information, structural correspondences, and contact regions. This design encourages the model to reconstruct the global structure and track transformations by using specific blocks as reference points. Specifically, we extend the three question types as follows (\cref{fig:spatialblock}(b)):

\begin{itemize}
    \item[\textbf{Q1:}] \textbf{3D-to-2D projection with depth cues} 
    To facilitate depth-aware reasoning, different colors are assigned based on depth from a given viewpoint. Thus, the task requires not only matching the silhouette but also tracking front–back relationships using color information. Color functions as an explicit indicator of depth ordering and is designed to clearly reflect the hierarchical organization of the 3D structure.
    
    \item[\textbf{Q2:}] \textbf{Viewpoint transformation with anchor blocks} 
    Among the blocks composing the 3D structure, one anchor block is assigned a distinct color. This block must maintain the same structural role before and after transformation. Accordingly, the task requires not only recognizing geometric structure but also understanding anchor-relative relationships that are preserved through transformation.
    
    \item[\textbf{Q3:}] \textbf{Structural combination with overlapping blocks} 
    The attachment location between two structures is represented not by arrows but by the overlap of colored blocks indicated by transparent regions. Each structure is presented as a separate image, so aligning the positions of same-colored blocks across images is needed to predict the combined structure. Color thus acts as a shared reference point linking two independent inputs, inducing reasoning inter-structural integration.
    
\end{itemize}

This visual cue–based extension is designed to maintain the requirement for understanding the 3D structure itself while additionally enabling the learning of scene organization and relational reasoning centered around anchor objects, as required in realistic visual contexts. Consequently, it leads to meaningful improvement in real-scene spatial reasoning performance, as empirically demonstrated in \cref{subsec:ab_vc}. Finally, the SpatialBlock-15k dataset comprises three question types with visual cue–based extensions, each containing 5k samples. Further details on data construction can be found in \cref{sec:app_data}.

\section{Method}
\label{sec:method}
To foster the spatial intelligence of LVLMs using SpatialBlock-15k, we propose two training strategies, targeting different aspects of the model's capabilities: (1) direct answer prediction, which focuses on cultivating rapid inference by mapping visual inputs directly to their corresponding answers (\cref{sec:method_sft}), and (2) reasoning-based prediction, which encourages the model to explain its intermediate logical paths that lead to a final answer (\cref{sec:method_rl}).

\subsection{Direct Answer Prediction Model}
\label{sec:method_sft}
To maximize the model's capacity for instantaneous spatial problem-solving, we train it with supervision only on the ground-truth answer sequence $y=\{y_1,y_2,...,y_T\}$. Specifically, given a textual query $q$ and image $v$, we optimize model parameters $\theta$ to predict the next token $y_i$, given the preceding context $y_{(1:i-1)}$, by minimizing standard cross-entropy loss: 
\begin{equation}
    \mathcal{L}_{\text{ce}}(\theta) = - \sum_{i} \log P \left( y_i \mid y_{(1:i-1)}, q, v \right).
\end{equation}
This strategy directly optimizes answer precision, thereby establishing a foundation for high-fidelity spatial problem-solving proficiency.

\subsection{Reasoning-based Prediction Model} 
\label{sec:method_rl}
To go beyond simple answer prediction, we adopt a second strategy that enables the model to articulate its internal logic. To ensure a solid foundation for spatial intelligence, we first initialize the model through supervised fine-tuning to establish basic task-solving ability, then employ reinforcement learning to facilitate high-order structural reasoning.

For model initialization, we train the model with Low-Rank Adaptation (LoRA) instead of full-parameter fine-tuning, as full fine-tuning often degrades the inherent reasoning ability. This lightweight adaptation maintains the model’s Chain-of-Thought (CoT) reasoning capabilities while aligning it with our dataset. Notably, we deviate from the conventional ``cold-start'' phase that relies on synthesized trajectories from larger teacher models~\citep{guo2025deepseek, li2026spatialladder}. The rationale behind this decision is that even state-of-the-art proprietary models exhibit relatively low accuracy on our dataset (\cref{tab:main_results}), making it infeasible to construct reliable high-quality trajectories. Our empirical analysis (\cref{subsec:ab_rl_init}) demonstrates a clear performance advantage in preserving the model's reasoning ability through LoRA-based tuning. This approach provides a more effective initialization for RL training than attempting to recover the thinking ability via a low-quality cold-start phase.

After LoRA-based initialization, we directly optimize the model using reinforcement learning via Group Relative Policy Optimization (GRPO). We design a multi-objective reward function that evaluates response correctness and reasoning trace quality with three components:
\begin{equation}
    \mathcal{R}(y,o) = \mathcal{R}_{\text{acc}}(y,o) + \mathcal{R}_{\text{format}}(o) + \mathcal{R}_{\text{len}}(o),
\end{equation}
where $o$ is a model-generated prediction given a question $c=(q,v)$ with ground-truth answer $y$. The accuracy reward $\mathcal{R}_{\text{acc}}$ verifies whether the final answer extracted from $o$ matches $y$. 

Next, the format reward $\mathcal{R}_{\text{format}}$ enforces the required CoT structure, with a comprehensive reasoning trace followed by the final answer enclosed within \texttt{<answer>} and \texttt{</answer>} tags. Lastly, to prevent overly brief or excessively verbose output, a length reward  $\mathcal{R}_{\text{len}}$ requires the response length $L$ to satisfy $50<L<1024$. For simplicity and stability, all components are implemented in binary rewards (0 or 1) based on whether each condition is satisfied.

Given this reward, for each question $c$, the old policy model $\pi_{\theta_{\text{old}}}$ samples a group of candidate responses $\{o_1,o_2,...,o_G\}$. 
% Each response $o_i$ receives a reward $r_i = \mathcal{R}(o, o_i)$, 
Each response $o_i$ receives a reward $r_i = \mathcal{R}(y, o_i)$,
from which we compute the advantage $A_i = \frac{r_i - \text{mean}(r_1, r_2, \dots, r_G)}{\text{std}(r_1, r_2, \dots, r_G)}$. The model is then updated by maximizing the following objective:
\begin{equation}
    \scriptstyle
    \mathcal{J}_{\text{GRPO}}(\theta) = \mathbb{E}_{c, \{o_i\}} \left[ \frac{1}{G} \sum_{i=1}^G \min \left( \frac{\pi_\theta(o_i|c)}{\pi_{\theta_{\text{old}}}(o_i|c)} A_i, \text{clip} \left( \frac{\pi_\theta(o_i|c)}{\pi_{\theta_{\text{old}}}(o_i|c)}, 1 \pm \varepsilon \right) A_i \right) - \beta \text{KL} [\pi_\theta || \pi_{\text{ref}}] \right],    
\end{equation}
where $\varepsilon$ and $\beta$ are hyper-parameters, and $\text{KL} [\pi_\theta || \pi_{\text{ref}}]$ is the KL divergence between the policy model $\pi_\theta$ and reference model $\pi_{\text{ref}}$.

\section{Experiments}
\label{sec:exp}
We conduct multiple experiments to validate the effectiveness of SpatialBlock-15k in fostering spatial intelligence. We first describe the experimental setup, including training details and evaluation protocols (\cref{sec:exp_settings}). We then compare our models with prior methods across several spatial reasoning benchmarks (\cref{sec:exp_main}). Next, we perform ablation studies to assess the effects of each component in dataset construction and training (\cref{sec:exp_ablation}). Lastly, we provide an analysis of the generalizability of our proposed dataset (\cref{sec:exp_anal}).

\subsection{Experimental Settings}
\label{sec:exp_settings}

We adopt Qwen2.5-VL-3B~\citep{bai2025qwen25vltechnicalreport}, Qwen2.5-VL-7B~\citep{bai2025qwen25vltechnicalreport}, Qwen3-VL-4B~\citep{bai2025qwen3}, and InternVL3-2B~\citep{zhu2025internvl3} as baseline models for our framework. To distinguish the training strategy, we append a suffix to the model name: \textit{SpatialBlock-direct} denotes the model that outputs the answer directly (\cref{sec:method_sft}), while \textit{SpatialBlock-reason} refers to the model that performs reasoning before producing the final answer (\cref{sec:method_rl}). Both variants are trained using the entire SpatialBlock-15k dataset.

To evaluate the spatial reasoning capability, we conduct experiments on five benchmarks spanning both in-domain and out-of-domain settings. For in-domain evaluation, we introduce SB-Bench, the held-out test split of SpatialBlock-15k, comprising 600 questions. For out-of-domain evaluation, we assess generalization to real-world scenarios on four benchmarks: 
MindCube~\citep{yin2025spatial}, MMSI-Bench~\citep{yang2026mmsi}, and SPBench~\citep{li2026spatialladder} for spatial reasoning, and MMMU~\citep{yue2024mmmu} for general visual perception. Across all benchmarks, we restrict evaluation to multiple-choice questions to align with our training answer format. Additional numerical results are provided in \cref{subsec:app_num}.

% MindCube~\citep{yin2025spatial} for multi-view mental modeling, MMSI-Bench~\citep{yang2026mmsi} for complex spatial reasoning in diverse scenes, SPBench~\citep{li2026spatialladder} for single- and multi-image spatial reasoning, and MMMU~\citep{yue2024mmmu} for general visual perception. Across all benchmarks, we restrict evaluation to multiple-choice questions, as our method is not designed for numerical prediction tasks.

\begin{table}[!t]
\centering
\small
\setlength{\tabcolsep}{6pt}
\caption{\textbf{Main Results}. For each dataset, \textbf{bold} and \underline{underlined} indicate the best and second-best performance, respectively. * represents in-domain results where the model was trained on the benchmark. We re-evaluated all previous works for fair comparison.} 
\label{tab:main_results}
\resizebox{\textwidth}{!}{
    \begin{tabular}{lcccccc>{\columncolor{lightgray}}c}
    \toprule
    \multirow{2}{*}{Model} & \multirow{2}{*}{\# Data} & {\textbf{In-domain}} 
    & \multicolumn{5}{c}{\textbf{Out-of-domain}} \\
    \cmidrule(lr){3-3} \cmidrule(lr){4-8}
     & & SB-Bench & MindCube & MMSI-Bench & SPBench & MMMU & Overall \\
    \midrule
    \midrule
    \rowcolor{skyblue}
    \multicolumn{8}{c}{\textit{\textbf{Proprietary Models}}} \\
    GPT-5~\citep{singh2025openai} & - & 43.8 & 56.7 & 42.8 & 43.0 & 77.6 & 55.0 \\
    Gemini-2.5-flash~\citep{comanici2025gemini} & - & 36.4 & 51.0 & 32.2 & 44.3 & 58.8 & 46.6 \\
    Claude-Sonnet-4.5~\citep{anthropic2025sonnet45card} & - & 46.3 & 34.0 & 32.7 & 43.9 & 69.2 & 43.9 \\

    \midrule
    \rowcolor{skyblue}
    \multicolumn{8}{c}{\textit{\textbf{Open-Source Models}}} \\
    LLaVA-OneVision-7B~\citep{li2024llava} & - & 23.8 & 39.0 & 28.4 & 43.4 & 42.6 & 39.2 \\
    InternVL3-8B~\citep{zhu2025internvl3} & - & 22.7 & 42.7 & 25.1 & 46.1 & 48.0 & 40.5 \\
    Llama-3.2-11B-Vision-Instruct~\citep{Dubey2024TheL3} & - & 23.0 & 29.5 & 24.5 & 28.2 & 40.7 & 30.7 \\

    \midrule
    \rowcolor{skyblue}
    \multicolumn{8}{c}{\textit{\textbf{Spatial Specialist Models}}} \\
    \rowcolor{dark_skyblue!10}
    \multicolumn{8}{c}{\textit{Qwen2.5-VL-3B Based Models}} \\
    Qwen2.5-VL-3B-Instruct~\citep{bai2025qwen25vltechnicalreport} & - & 29.9 & 39.1 & 26.1 & 40.0 & 45.6 & 37.7 \\
    SpaceQwen~\citep{chen2024spatialvlm} & 2B & 30.6 & 38.8 & 24.5 & \underline{40.8} & 45.6 & 37.4 \\
    SpatialMLLM~\citep{wu2025spatial} & 120K & 30.0 & 36.3 & 25.4 & \underline{40.8} & 40.4 & 35.7 \\
    % VST-3B-SFT &  & 25.7 & 40.3 & 28.6 & 50.4 & 49.4 & 42.2 \\
    % VST-3B-RL &  & 31.0 & 38.0 & 31.0 & 49.6 & 45.2 & 40.9 \\
    SpatialLadder-3B~\citep{li2026spatialladder} & 25K & 25.5 & \underline{46.4} & 26.0 & 76.6* & \textbf{47.3} & 39.9 \\
    Spatial-SSRL-3B~\citep{liu2026spatial} & 81K & 29.5 & 41.0 & 24.4 & 19.5 & 45.2 & 32.5 \\
    \textbf{SpatialBlock-3B-direct (Ours)} & 15K & \textbf{94.8} & \textbf{49.1} & \underline{28.1} & 40.4 & 46.4 & \underline{41.0} \\
    \textbf{SpatialBlock-3B-reason (Ours)} & 15K & \underline{90.2} & 45.3 & \textbf{29.2} & \textbf{43.7} & \underline{46.7} & \textbf{41.2} \\

    %\midrule
    \arrayrulecolor{gray}\midrule\arrayrulecolor{black}
    \rowcolor{dark_skyblue!10}
    \multicolumn{8}{c}{\textit{Qwen2.5-VL-7B Based Models}} \\
    Qwen2.5-VL-7B-Instruct~\citep{bai2025qwen25vltechnicalreport} & - & 20.8 & 31.3 & 27.4 & 46.7 & 46.6 & 38.0 \\
    % VST-7B-SFT &  & 31.8 & 36.4 & 31.9 & 56.4 & 52.7 & 44.4 \\
    % VST-7B-RL &  & 34.7 & 37.9 & 35.3 & 58.7 & 48.5 & 45.1 \\
    SpaceR~\citep{ouyang2025spacer} & 150K & 26.8 & 31.6 & 27.4 & \textbf{53.1} & \underline{55.4} & 41.9 \\
    Spatial-SSRL-7B~\citep{liu2026spatial} & 81K & 27.7 & 35.1 & 27.5 & 23.4 & 48.9 & 33.7 \\
    \textbf{SpatialBlock-7B-direct (Ours)} & 15K & \textbf{95.0} & \textbf{48.8} & \underline{28.4} & 48.7 & 45.2 & \underline{42.8} \\
    \textbf{SpatialBlock-7B-reason (Ours)} & 15K & \underline{77.5} & \underline{37.9} & \textbf{29.7} & \underline{50.9} & \textbf{55.5} & \textbf{43.5} \\

    %\midrule
    \arrayrulecolor{gray}\midrule\arrayrulecolor{black}
    \rowcolor{dark_skyblue!10}
    \multicolumn{8}{c}{\textit{Qwen3-VL-4B Based Models}} \\
    Qwen3-VL-4B-Instruct~\citep{bai2025qwen3} & - & 17.0 & 26.2 & 27.1 & \underline{44.5} & 47.5 & 36.3 \\
    Spatial-SSRL-4B~\citep{liu2026spatial} & 81K & 34.8 & 34.6 & \textbf{28.3} & 23.1 & \textbf{50.5} & 34.1 \\
    \textbf{SpatialBlock-4B-direct (Ours)} & 15K & \textbf{95.7} & \textbf{51.3} & \underline{27.8} & \textbf{45.9} & 47.2 & \textbf{43.1} \\
    \textbf{SpatialBlock-4B-reason (Ours)} & 15K & \underline{94.5} & \underline{41.1} & \textbf{28.3} & 43.7 & \underline{49.6} & \underline{40.7} \\

    \arrayrulecolor{gray}\midrule\arrayrulecolor{black}
    \rowcolor{dark_skyblue!10}
    \multicolumn{8}{c}{\textit{InternVL3-2B Based Models}} \\
    InternVL3-2B-Instruct~\citep{zhu2025internvl3} & - & 19.0 & 32.1 & 27.0 & 33.0 & 41.3 & 33.4 \\
    \textbf{SpatialBlock-2B-direct (Ours)} & 15K & \textbf{97.2} & \textbf{51.4} & \underline{27.3} & \underline{41.3} & \underline{42.4} & \textbf{40.6} \\
    \textbf{SpatialBlock-2B-reason (Ours)} & 15K & \underline{78.3} & \underline{38.9} & \textbf{27.8} & \textbf{45.6} & \textbf{47.3} & \underline{39.9} \\
    
    \bottomrule
    \end{tabular}

}
% \vspace{-0.7cm}
\end{table}

\subsection{Main Results}
\label{sec:exp_main}
\cref{tab:main_results} shows our method's performance on spatial reasoning benchmarks. Notably, despite being trained with only 15k synthetic SpatialBlock samples, our model consistently outperforms existing spatial specialists on real-scene benchmarks. This result highlights that our block-stacking formulation provides an effective training signal for spatial reasoning.

% 수치 언급만 있는 것 같아서 어떤 면이 좋아진 건지 언급 추가
For \textbf{SpatialBlock-direct} models, we observe substantial gains, particularly on MindCube, which requires mental simulation. Among Qwen-based models, the 3B model outperforms the state-of-the-art SpatialLadder by 2.7\%, while the 7B model shows an even larger gain of 17.6\% over its backbone. The 4B model further improves upon its backbone by 25.1\%, achieving 51.3\%, the best open-source performance. Beyond Qwen, our method also improves InternVL3, demonstrating its generalizability across model scales and families.
% For \textbf{SpatialBlock-direct} models, we observe substantial performance gains on MindCube, which requires mental simulation. Among the Qwen-based models, SpatialBlock-3B-direct outperforms the state-of-the-art SpatialLadder by 2.7\%, SpatialBlock-7B-direct and SpatialBlock-4B-direct surpass their respective backbones by 17.6\% and 25.1\%. Our method also improves InternVL3, demonstrating generalization beyond the Qwen family.

The reasoning-enhanced variant, \textbf{SpatialBlock-reason}, further demonstrates strong results on MMSI-Bench, which requires complex logical inference. The 3B model outperforms SpatialLadder-3B by 3.2\% using a simpler training pipeline based solely on synthetic block-stacking data, whereas the 7B model outperforms SpaceR and Spatial-SSRL with only 15K training samples. Beyond final-answer accuracy, our model also improves reasoning quality, achieving a higher alignment score of 21.1 with ground-truth reasoning steps, compared to 17.8 for the baseline. (See \cref{subsec:app_reval} for details.) On SPBench, our models consistently improve relative spatial reasoning despite such questions not being explicitly included in the training data, demonstrating strong generalization beyond the supervised task distribution.
%% SSRL 결과도 SpaceR 이랑 같이 적어놨어
%The 3B model outperforms SpatialLadder-3B by 3.2\% using a simpler training pipeline based solely on synthetic block-stacking data, whereas the 7B model outperforms SpaceR and Spatial-SSRL by about 2\% with only 15K training samples, versus 150K and 81K, respectively.
%The 3B model outperforms SpatialLadder-3B by 3.2\% using a simpler training pipeline based solely on synthetic block-stacking data, while the 7B model outperforms SpaceR by 2.3\% with only one-tenth the training data.
% The reasoning-enhanced variant, \textbf{SpatialBlock-reason}, further demonstrates strong results on MMSI-Bench, which requires complex logical inference. SpatialBlock-3B-reason outperforms SpatialLadder-3B by 3.2\% despite using a simpler training pipeline based solely on synthetic block-stacking data, while SpatialBlock-7B-reason outperforms SpaceR by 2.3\% with only one-tenth the training data size. To further analyze reasoning quality, we use GPT-5 to evaluate alignment with the ground-truth rationales of MMSI-Bench, scoring each answer as 0, 0.5, or 1 and scaling the total to 100. SpatialBlock-3B-reason achieves 21.1, compared to 17.8 for the baseline, indicating improved reasoning quality beyond formatting or instruction following.

In summary, direct models excel on benchmarks that require canonical 90-degree viewpoint transformations, while reasoning models perform better on tasks requiring diverse viewpoint changes and multi-image reasoning. Also, despite being trained without any real-scene images, our models maintain stable performance on the general visual understanding benchmark MMMU, indicating that SpatialBlock training enhances 3D spatial reasoning without degrading broader visual perception. 

\subsection{Ablation Studies}
\label{sec:exp_ablation}

\begin{table}[!t]
\centering
\small
\setlength{\tabcolsep}{6pt}
\caption{\textbf{Ablation Studies}. * denotes in-domain results from benchmark-trained models.}
\label{tab:ablation}
\resizebox{\textwidth}{!}{
    \begin{tabular}{lcccccc>{\columncolor{lightgray}}c}
    \toprule
    \multirow{2}{*}{Model} & \multirow{2}{*}{\# Data} & {\textbf{In-domain}} 
    & \multicolumn{5}{c}{\textbf{Out-of-domain}} \\
    \cmidrule(lr){3-3} \cmidrule(lr){4-8}
     & & SB-Bench & MindCube & MMSI-Bench & SPBench & MMMU & Overall \\
    \midrule
    \midrule

    Qwen2.5-VL-3B-Instruct~\citep{bai2025qwen25vltechnicalreport} & - & 29.9 & 39.1 & 26.1 & 40.0 & 45.6 & 37.7 \\
    SpatialBlock-3B-direct (Ours) & 15K & 94.8 & 49.1 & 28.1 & 40.4 & 46.4 & 41.0 \\
    SpatialBlock-3B-reason (Ours) & 15K & 90.2 & 45.3 & 29.2 & 43.7 & 46.7 & 41.2 \\

    \arrayrulecolor{gray}\midrule\arrayrulecolor{black}
    \rowcolor{dark_skyblue!10}
    \multicolumn{8}{c}{\textit{(a) Ablation on question types}} \\
    SpatialBlock-3B-reason (Q1 only) & 15K & 48.0 & 41.8 & 27.6 & 42.8 & 43.2 & 38.8 \\
    SpatialBlock-3B-reason (Q2 only) & 15K & 51.5 & 38.8 & 24.2 & 42.1 & 47.1 & 38.0 \\
    SpatialBlock-3B-reason (Q3 only) & 15K & 40.8 & 41.3 & 25.3 & 43.2 & 46.5 & 39.1 \\

    \arrayrulecolor{gray}\midrule\arrayrulecolor{black}
    \rowcolor{dark_skyblue!10}
    \multicolumn{8}{c}{\textit{(b) Ablation on visual cues}} \\
    SpatialBlock-3B-direct (w/o visual cues) & 15K & 61.7 & 41.2 & 26.2 & 38.4 & 46.6 & 38.1 \\
    SpatialBlock-3B-reason (w/o visual cues) & 15K & 55.8 & 38.0 & 24.4 & 44.8 & 48.8 & 39.0 \\

    \arrayrulecolor{gray}\midrule\arrayrulecolor{black}
    \rowcolor{dark_skyblue!10}
    \multicolumn{8}{c}{\textit{(c) Ablation on task design}} \\
    SpatialBlock-3B-direct (SpatialLadder-26k) & 15K & 24.7 & 44.5 & 27.2 & 72.1* & 45.3 & 47.3 \\
    SpatialBlock-3B-reason (SpatialLadder-26k) & 15K & 28.0 & 44.9 & 27.5 & 68.8* & 44.7 & 46.5 \\

    \arrayrulecolor{gray}\midrule\arrayrulecolor{black}
    \rowcolor{dark_skyblue!10}
    \multicolumn{8}{c}{\textit{(d) Ablation on model initialization}} \\
    SpatialBlock-3B-reason (No Init.) & 15K & 67.8 & 41.3 & 25.4 & 43.5 & 46.7 & 39.2 \\
    SpatialBlock-3B-reason (Full-SFT Init., Qwen2.5-VL-7B) & 15K & 96.5 & 43.0 & 27.6 & 47.8 & 44.7 & 40.8 \\
    SpatialBlock-3B-reason (Full-SFT Init., GPT-5) & 15K & 75.3 & 41.7 & 28.0 & 44.3 & 48.1 & 40.5 \\
    
    \bottomrule
    \end{tabular}

}
% \vspace{-0.4cm}
\end{table}

\subsubsection{Ablation on Question Types}
\label{subsec:ab_qtype}
We evaluate models trained on a single question type (\cref{tab:ablation}(a)). Each question type targets a distinct aspect of spatial reasoning: Q1 focuses on viewpoint changes, Q2 emphasizes spatial transformations such as 90-degree rotations, and Q3 requires integrating spatial information across multiple perspectives. The single-type variants exhibit different performance patterns across benchmarks, suggesting that each type provides distinct spatial supervision. However, while single-type variants show limited gains, training on all three types achieves the best performance across most benchmarks. These results indicate that the three types provide complementary supervision, and combining them leads to more robust spatial reasoning and generalization.

\subsubsection{Ablation on Visual Cues}
\label{subsec:ab_vc}
We introduce color manipulation as a visual cue to approximate how humans interpret real-world scenes using visually salient reference objects (\cref{subsec:dataset_color}). To evaluate its effectiveness, we compare models trained with and without visual cues. As shown in \cref{tab:ablation}(b), removing visual cues leads to performance drops in \textit{direct} and \textit{reason} models of 7.9\% and 7.3\% in MindCube, and 1.9\% and 4.8\% in MMSI-Bench, respectively. These results suggest that color cues help models better understand complex scenes and mimic human scene perception, improving spatial reasoning in real-world environments. 

\subsubsection{Ablation on Task Design}
\label{subsec:ab_task_design}
In order to analyze the effect of the task design itself more thoroughly, we train a model on SpatialLadder-26k using the identical training recipe. For a fair comparison, we use 11.5k multiple-choice questions matching our response format and train for 1.3 epochs to match the same training size. As shown in \cref{tab:ablation}(c), this model underperforms ours on most benchmarks except the in-domain benchmark SPBench, suggesting that our performance gains mainly stem from the proposed block-stacking task design.

\subsubsection{Ablation on model initialization for reasoning model}
\label{subsec:ab_rl_init}
To train a reasoning-capable model via reinforcement learning, we adopt a LoRA-based initialization strategy to enhance baseline spatial intelligence while preserving inherent reasoning ability. \cref{tab:ablation}(d) compares different initialization strategies and their effects on performance. Without prior fine-tuning, performance drops noticeably across benchmarks, indicating the importance of establishing basic spatial understanding. We further compare our approach with commonly used cold-start phases based on full-parameter tuning~\citep{guo2025deepseek, li2026spatialladder}. While cold-start training can improve instruction following and achieve higher accuracy convergence, generating reliable trajectories for our tasks is challenging: both a larger model (Qwen2.5-VL-7B) and a proprietary model (GPT-5) frequently produce inaccurate block descriptions and reasoning. Consequently, training with these noisy reasoning traces yields only marginal gains. In contrast, LoRA-based initialization provides substantial improvements, offering a strong starting point for reinforcement learning with both basic spatial intelligence and reliable instruction-following ability. 

\subsection{Generalizability of Block-Stacking Questions}
\label{sec:exp_anal}

\begin{figure}[!t]
    \centering
    \includegraphics[width=\linewidth]{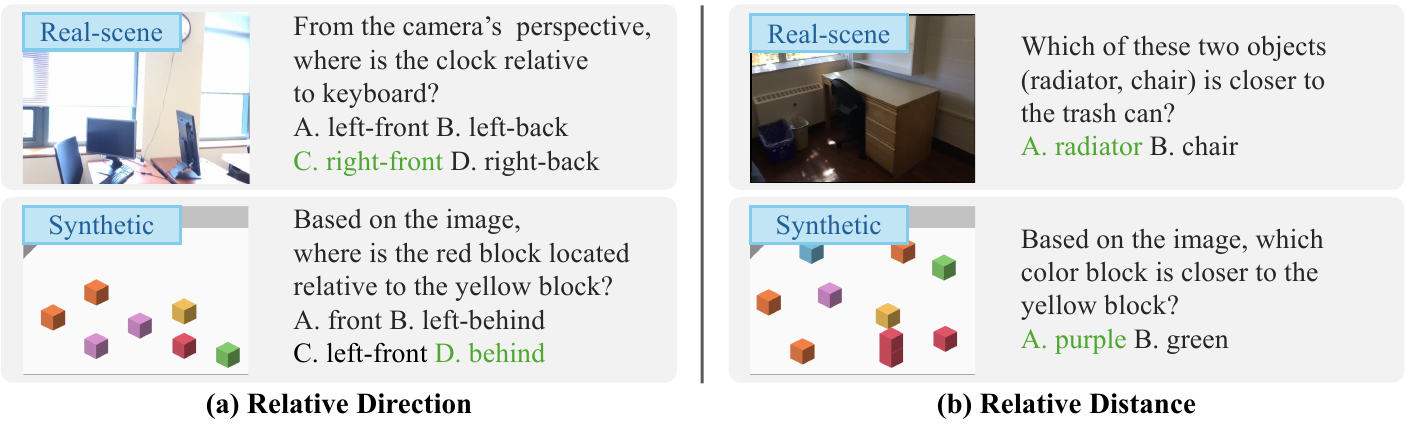}
    % \vspace{-0.8cm}
    \caption{\textbf{Synthetic-Real Dataset}. Widely used spatial question types, (a) relative direction and (b) relative distance, are adapted into a block-based environment while preserving their semantic structure.}
    \label{fig:synreal}
\end{figure}

\begin{table}[!t]
\centering
% \vspace{-0.4cm}
\setlength{\tabcolsep}{6pt}
\caption{\textbf{Comparison of Training on SpatialBlock and Synthetic-Real}. While Synthetic-Real improves in-domain performance, training on SpatialBlock leads to better generalization on out-of-domain benchmark datasets. All experiments are conducted on Qwen2.5-VL-3B~\citep{bai2025qwen25vltechnicalreport}, and in-domain performance is evaluated on SPBench.}
\label{tab:synreal}
\resizebox{\textwidth}{!}{
    \begin{tabular}{lccccc>{\columncolor{lightgray}}c}
    \toprule
    & \multicolumn{2}{c}{\textbf{In-domain}}
    & \multicolumn{3}{c}{\textbf{Out-of-domain}} \\
    \cmidrule(lr){2-3} \cmidrule(lr){4-6}
    Train Dataset
    & Relative Direction & Relative Distance & MindCube & MMSI-Bench & MMMU \\
    \midrule
    \midrule
    \textcolor{gray}{Baseline} & 31.4 & 74.1 & 39.1 & 26.1 & 45.6 \\
    Synthetic-Real & \textbf{43.4} & \textbf{76.9} & 32.8 & 25.5 & 43.6 \\
    SpatialBlock & 31.4 & 67.4 & \textbf{49.1} & \textbf{28.1} & \textbf{46.4} \\
    \bottomrule
    \end{tabular}
}
% \vspace{-0.4cm}
\end{table}

As shown in \cref{tab:main_results}, our models trained on SpatialBlock-15k generalize well to real-scene benchmark datasets despite involving question types that differ from those commonly used. This observation raises the question of whether our block-stacking tasks provide a more effective learning signal than conventional spatial question formulations. To investigate this, we construct \textit{Synthetic-Real}, a dataset of the same size as SpatialBlock-15k, which adapts two widely used spatial question types -- relative direction and distance -- into our block-based environment (\cref{fig:synreal}). Although both datasets can be generated with comparable effort, the results in \cref{tab:synreal} reveal that the model trained on Synthetic-Real achieves higher accuracy only on in-domain tasks and performs poorly on out-of-domain benchmark datasets. In contrast, our model consistently improves across benchmarks. These results suggest that SpatialBlock tasks promote a more holistic understanding of 3D scenes, leading to more generalizable and transferable spatial reasoning than standard spatial supervision.

\section{Conclusion}
\label{sec:conclusion}
In this paper, we revisited the problem of improving spatial intelligence in large vision-language models from a new perspective: learning foundational spatial skills through structured tasks rather than relying on annotation-heavy real-scene supervision. To this end, we introduced \textbf{SpatialBlock-15k}, a novel synthetic dataset of block-stacking problems, including 3D-to-2D projection, viewpoint transformation, and structural integration. The dataset further incorporated controlled color variation as visual cues to encourage anchor-based reasoning, mimicking how humans interpret scenes by using specific objects as reference points. Using this dataset, we developed two specialized models: \textbf{SpatialBlock-direct} for immediate answer generation and \textbf{SpatialBlock-reason} for detailed logical inference. Experimental results demonstrated that our models consistently outperform existing spatial specialists on various real-world spatial benchmarks. Furthermore, extensive analyses show that the learned capabilities generalize across tasks and domains, indicating that the models capture the underlying spatial structure of scenes. These findings highlighted that robust spatial intelligence can be achieved through training on carefully designed synthetic spatial tasks, offering a scalable and effective path for improving spatial reasoning in LVLMs.

% \subsubsection*{AI Use Statement}
% We used generative AI tools solely as a reference for improving the clarity and phrasing of the manuscript. The tools were not used to generate or draft substantive content, nor were they used for research ideation, methodology, experimental design, or interpretation of results. All final wording and content were reviewed and determined by the authors, who take full responsibility for the manuscript.

% \subsubsection*{Acknowledgments}

\bibliography{iclr2027_conference}
\bibliographystyle{iclr2027_conference}

\clearpage
\appendix
% \section{Appendix}
% \setcounter{page}{1}

\section{Details on Construction and Evaluation of SpatialBlock-15k}
\label{sec:app_data}

\subsection{Data Construction Process}
All 3D structures in SpatialBlock-15k are constructed on a $3\times3$ spatial grid, where each grid position can contain between one and four vertically stacked blocks. The height of each stack defines the underlying 3D configuration. To avoid ambiguity when interpreting rendered structures, we restrict the set of valid configurations so that the number of blocks at every grid position can be reliably inferred from the rendered image. In particular, configurations that have severe occlusion -- where blocks are completely hidden behind others from the viewing perspective -- are excluded. This constraint ensures that the global structure remains visually interpretable while still allowing a diverse set of configurations.

Based on the resulting valid structures, we generate the three types of tasks illustrated in \cref{fig:spatialblock}. For each task, the distractor choices are carefully constructed to be structurally plausible while incorrect, preventing simple heuristics and encouraging models to reason about the spatial relationships within the structure.

\subsubsection{Q1: 3D-to-2D projection with depth cues}
In this task, the model must predict the correct 2D projection (e.g., front or side view) of a given 3D block structure. The wrong choices are generated through controlled perturbations of the correct projection:

\begin{itemize}
\item \textbf{Block-removal perturbation.} A visible top block from one column is removed, producing a projection that corresponds to a slightly reduced stack height while remaining structurally plausible.
\item \textbf{Color-consistency perturbation.} For a column containing multiple colors, the color of one block is replaced with that of a neighboring block in the same column. This preserves the geometric structure but violates the correct color arrangement.
\item \textbf{Combined perturbation.} A hybrid modification that simultaneously removes a block from one column and alters the color assignment in another column.
\end{itemize}

During choice generation, perturbations are only applied to columns that contain visible blocks and sufficient color diversity, ensuring that distractors remain meaningful and visually distinguishable.

\subsubsection{Q2: Viewpoint transformation with anchor blocks}
This task evaluates whether models can reason about global spatial transformations of the structure. Given an original structure, one transformation is randomly selected from the following set: $90^\circ$, $180^\circ$, or $270^\circ$ rotations, and flipping vertically. A specific visible block in the original structure is highlighted, and the model needs to identify the corresponding block position after the transformation. To ensure valid supervision signals, only blocks that remain visible after transformation are selected as colored blocks.

The answer choices consist of four candidates:

\begin{itemize}
\item the correctly transformed structure with the correct mapped block position (correct answer),
\item the original structure with a different block highlighted,
\item the correctly transformed structure with an incorrectly highlighted block,
\item or a structure generated through a different transformation.
\end{itemize}

\subsubsection{Q3: Structural combination with overlapping blocks}
This task focuses on reasoning about the union when two block structures are combined. Each problem instance consists of two structures and a specified attachment point indicated by overlapping color. The goal is to determine the correct outcome of the combination. During dataset construction, we sample a possible combination from the given 3D structures and compute their first contact location as the overlap region.

The distractor choices are constructed by choosing a 3D configuration where one component is replaced, and the other remains unchanged, resulting in a different structure.
%one of the two interacting structures with a different structure while keeping the other structure unchanged. The resulting configuration is then recomputed using the same interaction rule. This process generates alternative outcomes that are geometrically plausible but inconsistent with the original pair of structures. 
By ensuring that each distractor differs only in one of the two input structures, the choices remain visually similar to the correct answer while requiring the model to reason about the precise spatial interaction between the specific pair of structures.

\subsection{Human Evaluation on SB-Bench}
\label{subsec:app_exp_human}
For human evaluation, we randomly select 30 samples from SB-Bench and collect responses from 31 participants. As illustrated in \cref{fig:intro}, humans achieve 95\% accuracy on average, demonstrating the task's simplicity.

\section{Training Implementation Details}
\label{sec:app_train_detail}

\subsection{Early stopping for SpatialBlock-direct}
To prevent overfitting to the block structure during direct model training, we selected the model from an intermediate iteration in which test accuracy exceeded 93\%, specifically the iteration immediately before it began to decrease.

\subsection{Reasoning prompt used for reinforcement learning}
\label{subsec:app_exp_prompt}

\begin{figure}[!ht]
    \centering
    \includegraphics[width=\linewidth]{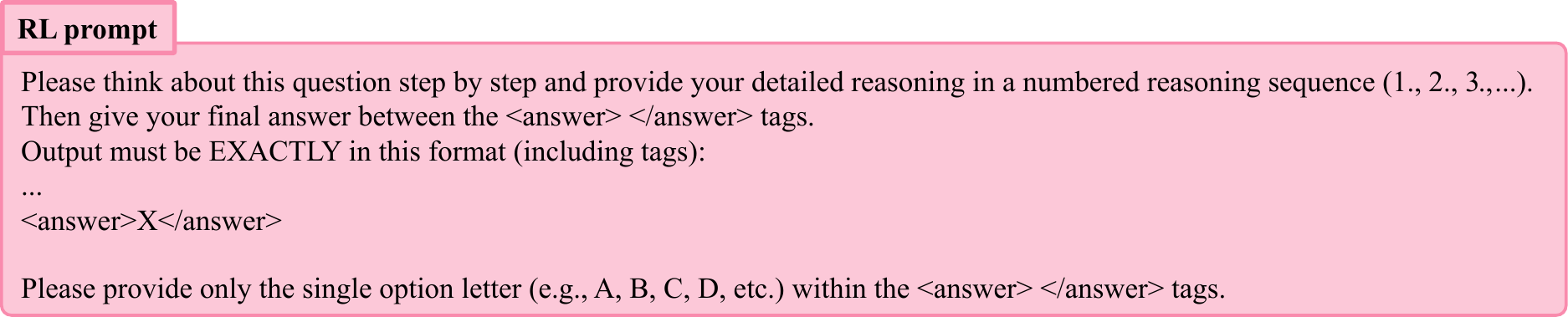}
    \caption{\textbf{Instructions for Reinforcement Learning}. The model is instructed to provide a step-by-step thinking process, and then the final answer between answer tags.}
    \label{fig:app_prompt}
\end{figure}

The detailed instructions for reinforcement learning are shown in \cref{fig:app_prompt}. The prompt guides the model to output reasoning steps in a numbered sequence and gives the final answer between \texttt{<answer>} and \texttt{</answer>} tags. We do not use the extra \texttt{<reason>} tag to reduce the load of learning the format.

\subsection{Hyperparameters used for training}
\label{subsec:app_hp}
We apply LoRA-based supervised fine-tuning for SpatialBlock-reason, using rank 8 for the 3B and 2B models and rank 16 for the 7B and 4B models. For reinforcement learning, we use EasyR1~\citep{zheng2025easyr1} as our base code and follow its default hyperparameters. We set the batch size to 512 with 5 rollouts, while extending the maximum prompt length to 4096 to accommodate multi-image training. All training was conducted for 1 epoch using SpatialBlock-15k.

\begin{figure}[!ht]
    \centering
    \includegraphics[width=\linewidth]{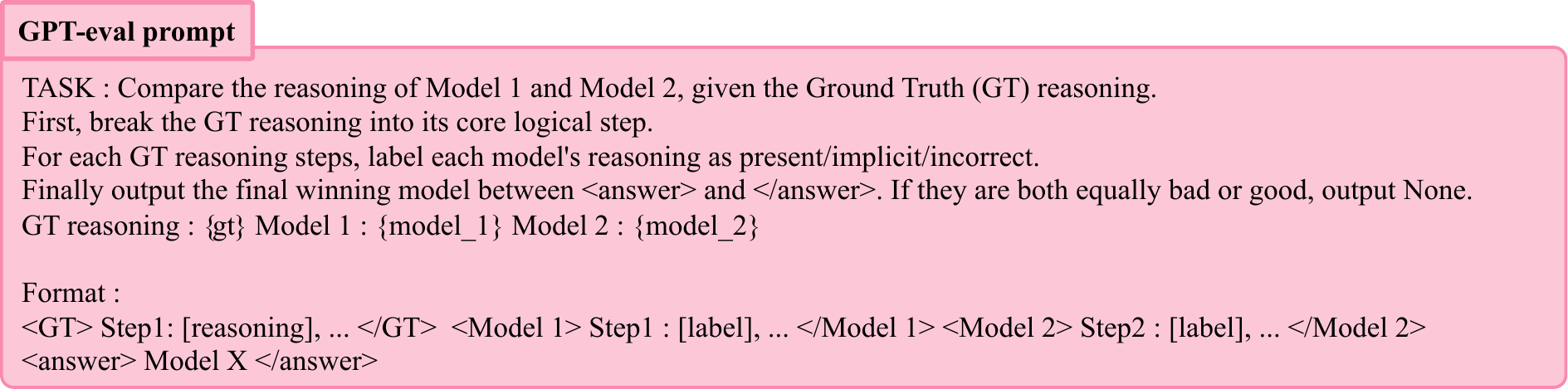}
    \caption{\textbf{Instructions for Reasoning Quality Evaluation}. We evaluate the quality of model-generated reasoning by measuring its alignment with each step of the corresponding ground-truth rationale.}
    \label{fig:app_reasoning_eval}
\end{figure}
% TODO) instruction

\subsection{Reasoning quality evaluation protocol}
\label{subsec:app_reval}
To further assess reasoning quality, we utilize MMSI-Bench with its ground-truth rationales. GPT-5 scores each reasoning's alignment with each ground-truth step as 1 (present), 0.5 (implicit), or 0 (incorrect), and the final score is averaged and scaled to 100. The detailed instructions are shown in \cref{fig:app_reasoning_eval}.

\section{Additional Results}
\label{sec:app_add_exp}

\subsection{Additional Results on Numerical Tasks}
\label{subsec:app_num}

\begin{table}[!ht]
    \caption{\textbf{Results on Numerical Tasks}. Results on tasks from SPBench.}
    \centering
    \resizebox{\linewidth}{!}{
    \begin{tabular}{lcccc}
    \toprule
    Model & Absolute distance & Object size & Object counting & Average \\
    \midrule
    \midrule
    Qwen2.5-VL-3B-Instruct & 30.9 & 17.6 & 35.1 & 27.9 \\
    SpatialBlock-3B-direct & 23.2 & 19.9 & 72.1 & 38.4 \\
    SpatialBlock-3B-reason & 21.4 & 38.3 & 43.3 & 34.3 \\
    \bottomrule
    \end{tabular}
    }
    \label{tab:app_num}
\end{table}

We evaluate our model on numerical prediction tasks from SPBench (\cref{tab:app_num}). Despite being trained only on multiple-choice questions, our model improves in counting and size estimation, as our dataset enhances the model's ability to individuate objects and infer their scales. In contrast, it is less effective at absolute distance because our method prioritizes structural understanding and spatial integration over precise metric regression. 

\subsection{Robustness to viewing angle variations}
\label{subsec:app_robust}

\begin{figure}[!ht]
    \centering
    \includegraphics[width=\linewidth]{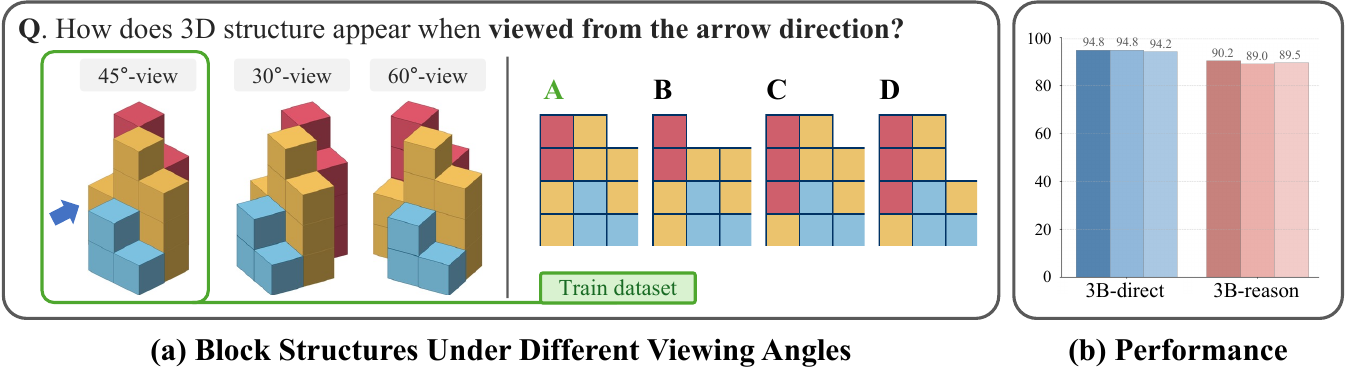}
    % \vspace{-0.8cm}
    \caption{\textbf{Performance Across Viewing Angles}. (a) The same block structures are rendered from different viewing angles. (b) Performance of models across viewing angles; the bars correspond to results under $45^{\circ}$, $30^{\circ}$, and $60^{\circ}$ views, respectively.}
    \label{fig:robustness}
\end{figure}

To further evaluate whether the model learns the underlying 3D structure rather than memorizing appearance patterns from a fixed viewpoint, we conduct a viewpoint generalization experiment. Our SpatialBlock-15k dataset is constructed using blocks rendered from a $45^{\circ}$ viewing angle, which is fixed during training. We generate the same problems using identical block configurations rendered from different viewing angles ($30^{\circ}$ and $60^{\circ}$) (\cref{fig:robustness}(a)). As shown in \cref{fig:robustness}(b), both the direct answering and reasoning-based models maintain consistent performance across these viewpoints. This suggests that the models capture the underlying structure of the blocks, demonstrating genuine enhancement in spatial ability.

\subsection{Cold-start data quality}
\label{subsec:app_cs_quality}
\begin{figure}[!ht]
    \centering
    \includegraphics[width=\linewidth]{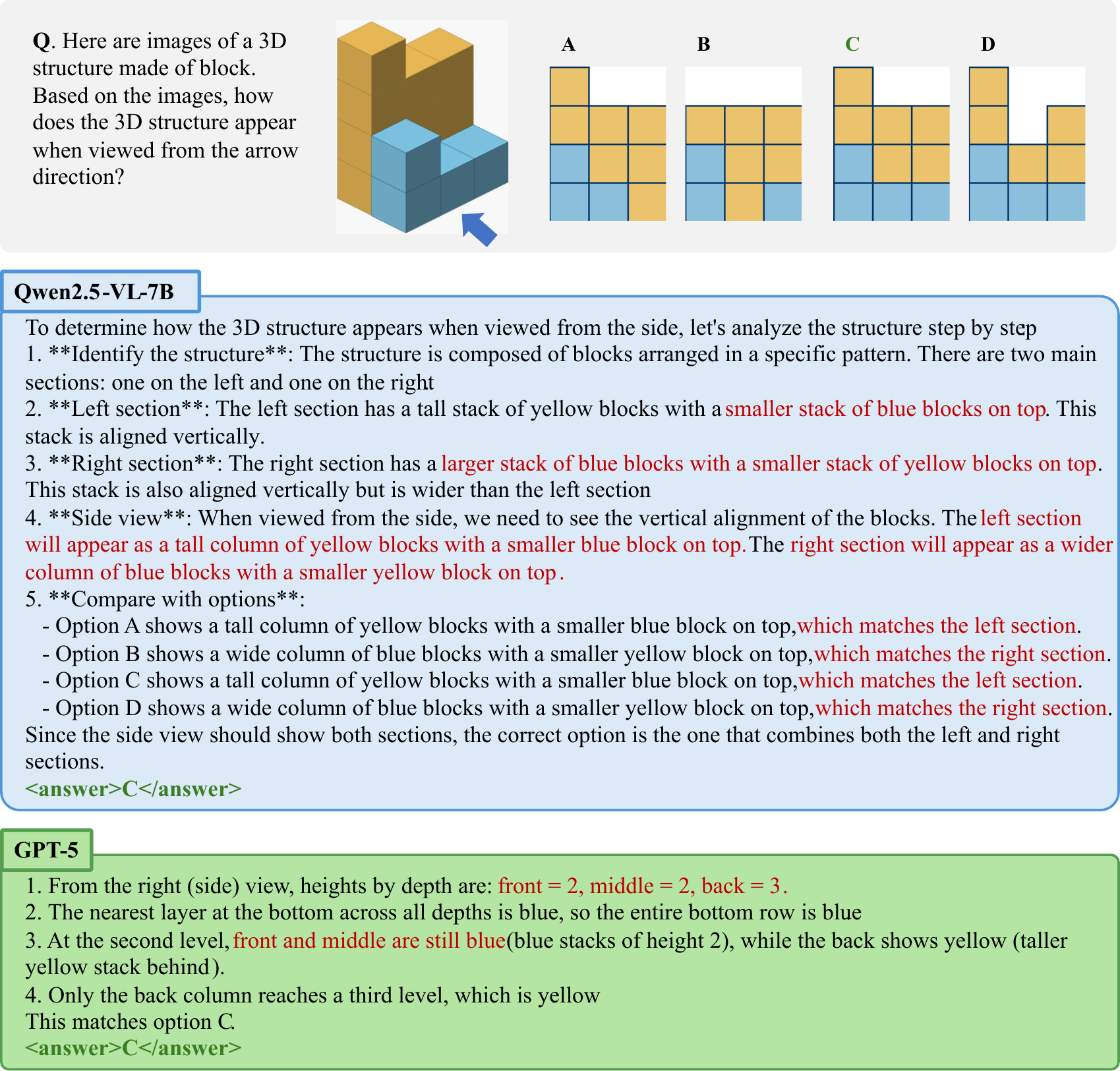}
    \caption{\textbf{Generated Cold-Start Data Sample}. The reasoning trace extracted from Qwen2.5-VL-7B includes incorrect information about the given 3D structure, even though the model answers correctly.}
    \label{fig:app_cs}
\end{figure}

As mentioned in \cref{subsec:ab_rl_init}, full-parameter fine-tuning with a cold-start phase shows marginal improvement compared with LoRA-based initialization. We hypothesize that this is due to the low quality of the cold-start reasoning trace. \cref{fig:app_cs} shows a reasoning example extracted from Qwen2.5-VL-7B. Although we choose the one with correct answers, the model fails to explain the 3D structure correctly, particularly miscounting the blocks. Furthermore, despite the attempt to evaluate all options, it generates the same explanation for distinct options, leading to a random final choice. 

We also observed that even proprietary models such as GPT-5 frequently generate inaccurate block descriptions and reasoning (\cref{fig:app_cs}), highlighting the difficulty of producing stable trajectories even for top-tier models on our tasks. Using these GPT-5-generated labels for cold-start SFT still underperforms ours, especially on spatial reasoning benchmarks (\cref{tab:ablation}(d)), which supports our claim that low-quality teacher reasoning can serve as noisy supervision. 
% TODO) GPT-5 cold-start figure 추가 (위 figure에 추가하면 좋을 듯? 같은 question에 대해 좀 더 나아진 reasoning이지만 여전히 틀린)

\clearpage
\subsection{Qualitative Result}
\label{subsec:app_add_qual}

\begin{figure}[!ht]
    \centering
    \includegraphics[width=\linewidth]{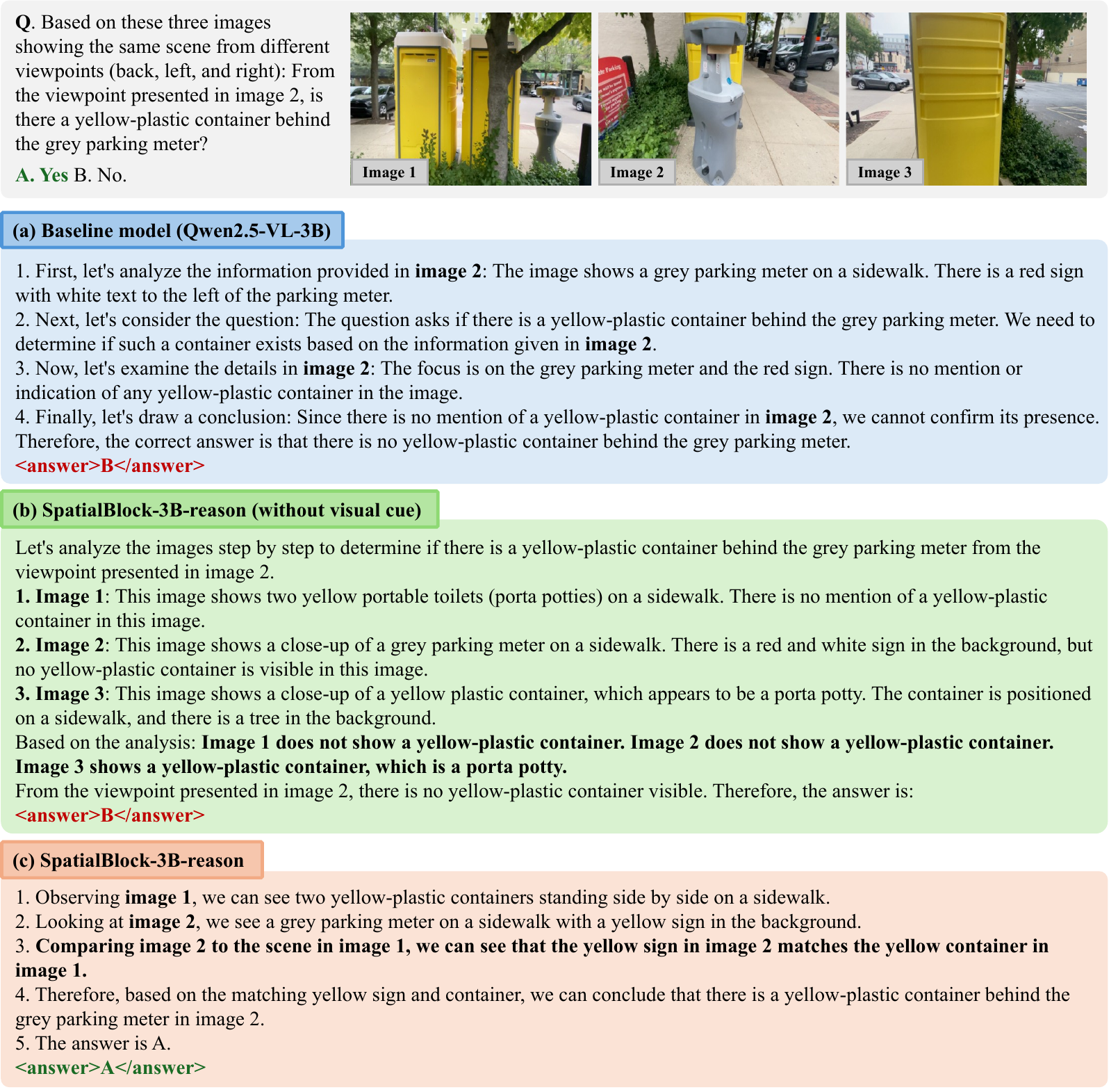}
    % \vspace{-0.7cm}
    \caption{\textbf{Qualitative Comparison of Reasoning Traces}. (a) Baseline only describes a single view. (b) Training without visual cues attempts multi-image reasoning but fails to identify the shared object. (c) Our model uses the ``yellow container'' as an anchor to integrate views and reach the correct answer.}
    \label{fig:qualitative}
\end{figure}

\cref{fig:qualitative} shows reasoning traces of the baseline model (Qwen2.5-VL-3B) and models trained on SpatialBlock-15k with and without visual cues for a question from the MindCube dataset. The task requires aggregating visual information from three images captured from different viewpoints. In addition, the model should infer the spatial relationship between two objects from the perspective presented in the second image (image 2). 

As shown in \cref{fig:qualitative}(a), the baseline model fails to exhibit multi-image understanding and merely describes image 2. The model trained without visual cues (\cref{fig:qualitative}(b)) attempts to reason over multiple images, but fails to identify the common object across views, leading to an incorrect conclusion. In contrast, SpatialBlock-3B-reason (\cref{fig:qualitative}(c)) identifies the ``yellow container'' as an anchor object and connects the given images through this shared reference, ultimately producing the correct answer. 

\begin{figure}[!ht]
    \centering
    \includegraphics[width=\linewidth]{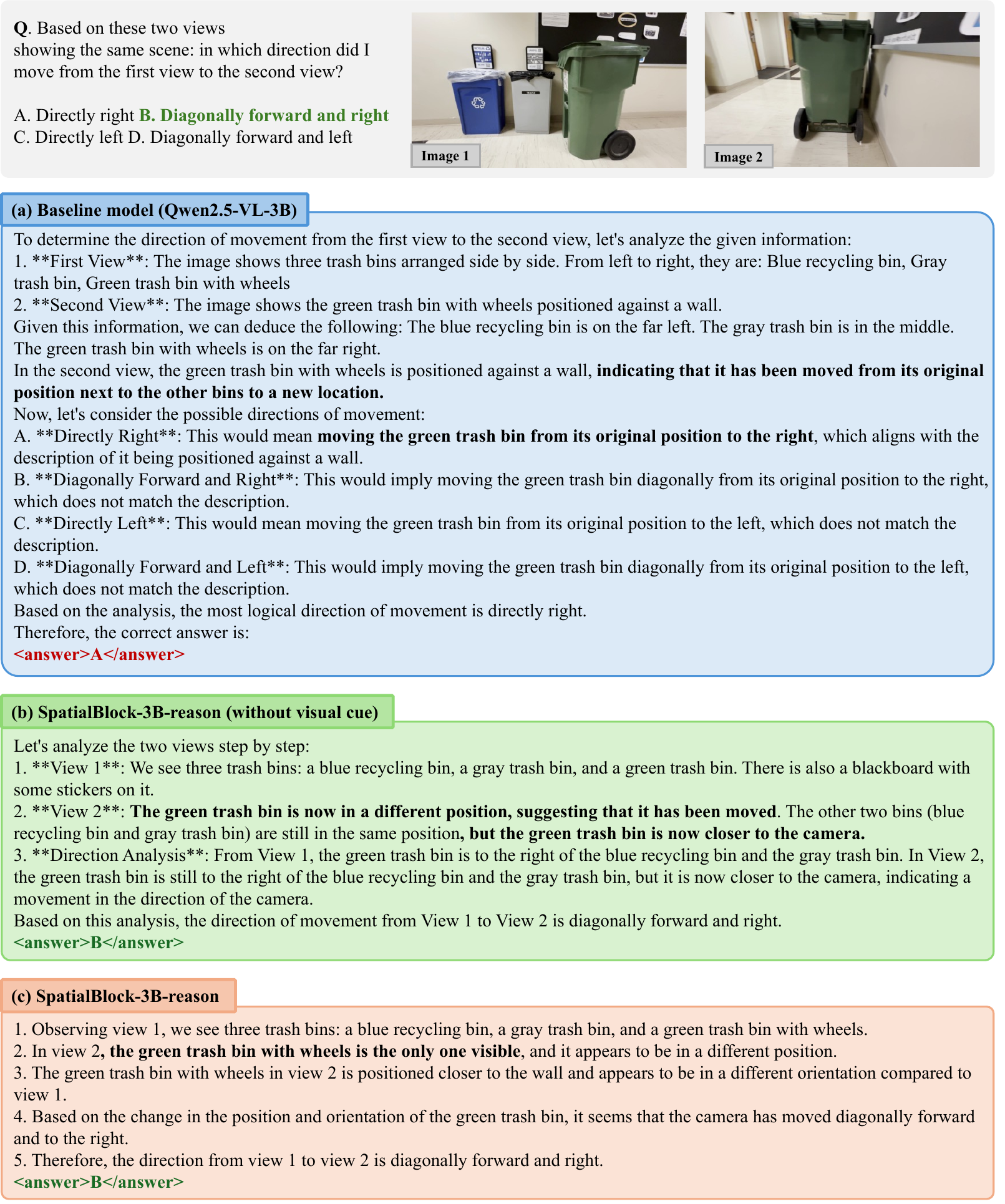}
    % \vspace{-0.7cm}
    \caption{\textbf{Additional Qualitative Result on Reasoning Traces}. The figure shows additional qualitative results on reasoning traces of the baseline model, our model trained without visual cues, and our final model, SpatialBlock-3B-reason.}
    \label{fig:app_add_qual}
\end{figure}

These results demonstrate that training on our dataset enhances the model's ability to perform complex spatial reasoning. Furthermore, incorporating visual cues enables the model to effectively leverage anchor objects to integrate information across multiple viewpoints. Additional qualitative results are illustrated in \cref{fig:app_add_qual}.

\end{document}

%% file: math_commands.tex
\usepackage{amsmath,amsfonts,bm}

\def\eqref#1{equation~\ref{#1}}
\def\1{\bm{1}}

\DeclareMathAlphabet{\mathsfit}{\encodingdefault}{\sfdefault}{m}{sl}
\SetMathAlphabet{\mathsfit}{bold}{\encodingdefault}{\sfdefault}{bx}{n}

%% file: iclr2027_conference.bib
@String(ICCV  = {Int. Conf. Comput. Vis.})

@String(ICCV  = {ICCV})

@article{levine2012early,
  title={Early puzzle play: a predictor of preschoolers' spatial transformation skill.},
  author={Levine, Susan C and Ratliff, Kristin R and Huttenlocher, Janellen and Cannon, Joanna},
  journal={Developmental psychology},
  volume={48},
  number={2},
  pages={530},
  year={2012},
  publisher={American Psychological Association}
}

@article{jirout2015building,
  title={Building blocks for developing spatial skills: Evidence from a large, representative US sample},
  author={Jirout, Jamie J and Newcombe, Nora S},
  journal={Psychological science},
  volume={26},
  number={3},
  pages={302--310},
  year={2015},
  publisher={Sage Publications Sage CA: Los Angeles, CA}
}

@article{caldera1999children,
  title={Children’s play preferences, construction play with blocks, and visual-spatial skills: Are they related?},
  author={Caldera, Yvonne M and Culp, Anne McDonald and O’Brien, Marion and Truglio, Rosemarie T and Alvarez, Mildred and Huston, Aletha C},
  journal={International Journal of Behavioral Development},
  volume={23},
  number={4},
  pages={855--872},
  year={1999},
  publisher={Sage Publications Sage CA: Thousand Oaks, CA}
}

@article{casey2008development,
  title={The development of spatial skills through interventions involving block building activities},
  author={Casey, Beth M and Andrews, Nicole and Schindler, Holly and Kersh, Joanne E and Samper, Alexandra and Copley, Juanita},
  journal={Cognition and instruction},
  volume={26},
  number={3},
  pages={269--309},
  year={2008},
  publisher={Taylor \& Francis}
}

@article{land2001ways,
  title={In what ways do eye movements contribute to everyday activities?},
  author={Land, Michael F and Hayhoe, Mary},
  journal={Vision research},
  volume={41},
  number={25-26},
  pages={3559--3565},
  year={2001},
  publisher={Elsevier}
}

@article{tong2024cambrian,
  title={Cambrian-1: A fully open, vision-centric exploration of multimodal llms},
  author={Tong, Peter and Brown, Ellis and Wu, Penghao and Woo, Sanghyun and Iyer, Adithya Jairam Vedagiri and Akula, Sai Charitha and Yang, Shusheng and Yang, Jihan and Middepogu, Manoj and Wang, Ziteng and others},
  journal={Advances in Neural Information Processing Systems},
  volume={37},
  pages={87310--87356},
  year={2024}
}

@inproceedings{lou2025llava,
  title={LLaVA-SP: Enhancing Visual Representation with Visual Spatial Tokens for MLLMs},
  author={Lou, Haoran and Fan, Chunxiao and Liu, Ziyan and Wu, Yuexin and Wang, Xinliang},
  booktitle={Proceedings of the IEEE/CVF International Conference on Computer Vision},
  pages={22014--22024},
  year={2025}
}

@article{cheng2024spatialrgpt,
  title={Spatialrgpt: Grounded spatial reasoning in vision-language models},
  author={Cheng, An-Chieh and Yin, Hongxu and Fu, Yang and Guo, Qiushan and Yang, Ruihan and Kautz, Jan and Wang, Xiaolong and Liu, Sifei},
  journal={Advances in Neural Information Processing Systems},
  volume={37},
  pages={135062--135093},
  year={2024}
}

@article{wu2025spatial,
  title={Spatial-mllm: Boosting mllm capabilities in visual-based spatial intelligence},
  author={Wu, Diankun and Liu, Fangfu and Hung, Yi-Hsin and Duan, Yueqi},
  journal={arXiv preprint arXiv:2505.23747},
  year={2025}
}

@inproceedings{li2026spatialladder,
  title={Spatialladder: Progressive training for spatial reasoning in vision-language models},
  author={Li, Hongxing and Li, Dingming and Wang, Zixuan and Yan, Yuchen and Wu, Hang and Zhang, Wenqi and Shen, Yongliang and Lu, Weiming and Xiao, Jun and Zhuang, Yueting},
  booktitle={International Conference on Learning Representations},
  volume={2026},
  pages={76566--76592},
  year={2026}
}

@article{ma2025spatialreasoner,
  title={Spatialreasoner: Towards explicit and generalizable 3d spatial reasoning},
  author={Ma, Wufei and Chou, Yu-Cheng and Liu, Qihao and Wang, Xingrui and de Melo, Celso and Xie, Jianwen and Yuille, Alan},
  journal={arXiv preprint arXiv:2504.20024},
  year={2025}
}

@inproceedings{yang2025thinking,
  title={Thinking in space: How multimodal large language models see, remember, and recall spaces},
  author={Yang, Jihan and Yang, Shusheng and Gupta, Anjali W and Han, Rilyn and Fei-Fei, Li and Xie, Saining},
  booktitle={Proceedings of the Computer Vision and Pattern Recognition Conference},
  pages={10632--10643},
  year={2025}
}

@inproceedings{yin2025spatial,
  title={Spatial mental modeling from limited views},
  author={Yin, Baiqiao and Wang, Qineng and Zhang, Pingyue and Zhang, Jianshu and Wang, Kangrui and Wang, Zihan and Zhang, Jieyu and Chandrasegaran, Keshigeyan and Liu, Han and Krishna, Ranjay and others},
  booktitle={Structural Priors for Vision Workshop at ICCV'25},
  year={2025}
}

@article{ouyang2025spacer,
  title={Spacer: Reinforcing mllms in video spatial reasoning},
  author={Ouyang, Kun and Liu, Yuanxin and Wu, Haoning and Liu, Yi and Zhou, Hao and Zhou, Jie and Meng, Fandong and Sun, Xu},
  journal={arXiv preprint arXiv:2504.01805},
  year={2025}
}

@inproceedings{chen2024spatialvlm,
  title={Spatialvlm: Endowing vision-language models with spatial reasoning capabilities},
  author={Chen, Boyuan and Xu, Zhuo and Kirmani, Sean and Ichter, Brain and Sadigh, Dorsa and Guibas, Leonidas and Xia, Fei},
  booktitle={Proceedings of the IEEE/CVF Conference on Computer Vision and Pattern Recognition},
  pages={14455--14465},
  year={2024}
}

@inproceedings{ma2025spatialllm,
  title={Spatialllm: A compound 3d-informed design towards spatially-intelligent large multimodal models},
  author={Ma, Wufei and Ye, Luoxin and de Melo, Celso M and Yuille, Alan and Chen, Jieneng},
  booktitle={Proceedings of the Computer Vision and Pattern Recognition Conference},
  pages={17249--17260},
  year={2025}
}

@article{wang2024qwen2,
  title={Qwen2-vl: Enhancing vision-language model's perception of the world at any resolution},
  author={Wang, Peng and Bai, Shuai and Tan, Sinan and Wang, Shijie and Fan, Zhihao and Bai, Jinze and Chen, Keqin and Liu, Xuejing and Wang, Jialin and Ge, Wenbin and others},
  journal={arXiv preprint arXiv:2409.12191},
  year={2024}
}

@inproceedings{tian2025nuscenes,
  title={Nuscenes-spatialqa: A spatial understanding and reasoning benchmark for vision-language models in autonomous driving},
  author={Tian, Kexin and Mao, Jingrui and Zhang, Yunlong and Jiang, Jiwan and Zhou, Yang and Tu, Zhengzhong},
  booktitle={Proceedings of the IEEE/CVF International Conference on Computer Vision},
  pages={4567--4576},
  year={2025}
}

@article{feng2025seeing,
  title={Seeing across views: Benchmarking spatial reasoning of vision-language models in robotic scenes},
  author={Feng, Zhiyuan and Kang, Zhaolu and Wang, Qijie and Du, Zhiying and Yan, Jiongrui and Shi, Shubin and Yuan, Chengbo and Liang, Huizhi and Deng, Yu and Li, Qixiu and others},
  journal={arXiv preprint arXiv:2510.19400},
  year={2025}
}

@article{guo2025deepseek,
  title={Deepseek-r1: Incentivizing reasoning capability in llms via reinforcement learning},
  author={Guo, Daya and Yang, Dejian and Zhang, Haowei and Song, Junxiao and Wang, Peiyi and Zhu, Qihao and Xu, Runxin and Zhang, Ruoyu and Ma, Shirong and Bi, Xiao and others},
  journal={arXiv preprint arXiv:2501.12948},
  year={2025}
}

@article{singh2025openai,
  title={Openai gpt-5 system card},
  author={Singh, Aaditya and Fry, Adam and Perelman, Adam and Tart, Adam and Ganesh, Adi and El-Kishky, Ahmed and McLaughlin, Aidan and Low, Aiden and Ostrow, AJ and Ananthram, Akhila and others},
  journal={arXiv preprint arXiv:2601.03267},
  year={2025}
}

@article{comanici2025gemini,
  title={Gemini 2.5: Pushing the frontier with advanced reasoning, multimodality, long context, and next generation agentic capabilities},
  author={Comanici, Gheorghe and Bieber, Eric and Schaekermann, Mike and Pasupat, Ice and Sachdeva, Noveen and Dhillon, Inderjit and Blistein, Marcel and Ram, Ori and Zhang, Dan and Rosen, Evan and others},
  journal={arXiv preprint arXiv:2507.06261},
  year={2025}
}

@article{li2024llava,
  title={Llava-onevision: Easy visual task transfer},
  author={Li, Bo and Zhang, Yuanhan and Guo, Dong and Zhang, Renrui and Li, Feng and Zhang, Hao and Zhang, Kaichen and Zhang, Peiyuan and Li, Yanwei and Liu, Ziwei and others},
  journal={arXiv preprint arXiv:2408.03326},
  year={2024}
}

@article{zhu2025internvl3,
  title={Internvl3: Exploring advanced training and test-time recipes for open-source multimodal models},
  author={Zhu, Jinguo and Wang, Weiyun and Chen, Zhe and Liu, Zhaoyang and Ye, Shenglong and Gu, Lixin and Tian, Hao and Duan, Yuchen and Su, Weijie and Shao, Jie and others},
  journal={arXiv preprint arXiv:2504.10479},
  year={2025}
}

@article{bai2025qwen3,
  title={Qwen3-vl technical report},
  author={Bai, Shuai and Cai, Yuxuan and Chen, Ruizhe and Chen, Keqin and Chen, Xionghui and Cheng, Zesen and Deng, Lianghao and Ding, Wei and Gao, Chang and Ge, Chunjiang and others},
  journal={arXiv preprint arXiv:2511.21631},
  year={2025}
}

@misc{bai2025qwen25vltechnicalreport,
      title={Qwen2.5-VL Technical Report}, 
      author={Shuai Bai and Keqin Chen and Xuejing Liu and Jialin Wang and Wenbin Ge and Sibo Song and Kai Dang and Peng Wang and Shijie Wang and Jun Tang and Humen Zhong and Yuanzhi Zhu and Mingkun Yang and Zhaohai Li and Jianqiang Wan and Pengfei Wang and Wei Ding and Zheren Fu and Yiheng Xu and Jiabo Ye and Xi Zhang and Tianbao Xie and Zesen Cheng and Hang Zhang and Zhibo Yang and Haiyang Xu and Junyang Lin},
      year={2025},
      eprint={2502.13923},
      archivePrefix={arXiv},
      primaryClass={cs.CV},
      url={https://arxiv.org/abs/2502.13923}, 
}

@inproceedings{yang2026mmsi,
  title={Mmsi-bench: A benchmark for multi-image spatial intelligence},
  author={Yang, Sihan and Xu, Runsen and Xie, Yiman and Yang, Sizhe and Li, Mo and Lin, Jingli and Zhu, Chenming and Chen, Xiaochen and Duan, Haodong and Yue, Xiangyu and others},
  booktitle={International Conference on Learning Representations},
  volume={2026},
  pages={157051--157088},
  year={2026}
}

@inproceedings{yue2024mmmu,
  title={Mmmu: A massive multi-discipline multimodal understanding and reasoning benchmark for expert agi},
  author={Yue, Xiang and Ni, Yuansheng and Zhang, Kai and Zheng, Tianyu and Liu, Ruoqi and Zhang, Ge and Stevens, Samuel and Jiang, Dongfu and Ren, Weiming and Sun, Yuxuan and others},
  booktitle={Proceedings of the IEEE/CVF conference on computer vision and pattern recognition},
  pages={9556--9567},
  year={2024}
}

@misc{zheng2025easyr1,
  title        = {EasyR1: An Efficient, Scalable, Multi-Modality RL Training Framework},
  author       = {Yaowei Zheng, Junting Lu and Shenzhi Wang, Zhangchi Feng and Dongdong Kuang, Yuwen Xiong and Richong Zhang},
  howpublished = {\url{https://github.com/hiyouga/EasyR1}},
  year         = {2025}
}

@techreport{anthropic2025sonnet45card,
  title       = {Claude Sonnet 4.5 System Card},
  author      = {Anthropic},
  institution = {Anthropic},
  year        = {2025},
  url         = {https://assets.anthropic.com/m/12f214efcc2f457a/original/Claude-Sonnet-4-5-System-Card.pdf},
}

@misc{Dubey2024TheL3,
  title={The Llama 3 Herd of Models},
  author={Abhimanyu Dubey and Abhinav Jauhri and Abhinav Pandey and Abhishek Kadian and Ahmad Al-Dahle and Aiesha Letman and Akhil Mathur and Alan Schelten and Amy Yang and Angela Fan and Anirudh Goyal and Anthony S. Hartshorn and Aobo Yang and Archi Mitra and Archie Sravankumar and Artem Korenev and Arthur Hinsvark and Arun Rao and Aston Zhang and Aur'elien Rodriguez and Austen Gregerson and Ava Spataru and Baptiste Rozi{\`e}re and Bethany M. Biron and Binh Tang and Bobbie Chern and Char-lotte Caucheteux and Chaya Nayak and Chloe Bi and Chris Marra and Chris McConnell and Christian Keller and Christophe Touret and Chunyang Wu and Corinne Wong and Cristian Canton Ferrer and Cyrus Nikolaidis and Damien Allonsius and Daniel J. Song and Danielle Pintz and Danny Livshits and David Esiobu and Dhruv Choudhary and Dhruv Mahajan and Diego Garcia-Olano and Diego Perino and Dieuwke Hupkes and Egor Lakomkin and Ehab A. AlBadawy and E I Lobanova and Emily Dinan and Eric Michael Smith and Filip Radenovic and Frank Zhang and Gabriel Synnaeve and Gabrielle Lee and Georgia Lewis Anderson and Graeme Nail and Gr{\'e}goire Mialon and Guanglong Pang and Guillem Cu-curell and Hailey Nguyen and Hannah Korevaar and Hu Xu and Hugo Touvron and Iliyan Zarov and Imanol Arrieta Ibarra and Isabel M. Kloumann and Ishan Misra and Ivan Evtimov and Jade Copet and Jaewon Lee and Jan Geffert and Jana Vranes and Jason Park and Jay Mahadeokar and Jeet Shah and Jelmer van der Linde and Jennifer Billock and Jenny Hong and Jenya Lee and Jeremy Fu and Jianfeng Chi and Jianyu Huang and Jiawen Liu and Jie Wang and Jiecao Yu and Joanna Bitton and Joe Spisak and Jongsoo Park and Joseph Rocca and Joshua Johnstun and Joshua Saxe and Ju-Qing Jia and Kalyan Vasuden Alwala and K. Upasani and Kate Plawiak and Keqian Li and Kenneth Heafield and Kevin R. Stone and Khalid El-Arini and Krithika Iyer and Kshitiz Malik and Kuen-ley Chiu and Kunal Bhalla and Lauren Rantala-Yeary and Laurens van der Maaten and Lawrence Chen and Liang Tan and Liz Jenkins and Louis Martin and Lovish Madaan and Lubo Malo and Lukas Blecher and Lukas Landzaat and Luke de Oliveira and Madeline Muzzi and Ma-hesh Pasupuleti and Mannat Singh and Manohar Paluri and Marcin Kardas and Mathew Oldham and Mathieu Rita and Maya Pavlova and Melissa Hall Melanie Kambadur and Mike Lewis and Min Si and Mitesh Kumar Singh and Mona Hassan and Naman Goyal and Narjes Torabi and Nikolay Bash-lykov and Nikolay Bogoychev and Niladri S. Chatterji and Olivier Duchenne and Onur cCelebi and Patrick Alrassy and Pengchuan Zhang and Pengwei Li and Petar Vasi{\'c} and Peter Weng and Prajjwal Bhargava and Pratik Dubal and Praveen Krishnan and Punit Singh Koura and Puxin Xu and Qing He and Qingxiao Dong and R. S. Mughil Srinivasan and Raj Ganapathy and Ramon Calderer and Ricardo Silveira Cabral and Robert Stojnic and Roberta Raileanu and Rohit Girdhar and Rohit Patel and Romain Sauvestre and Ron-nie Polidoro and Roshan Sumbaly and Ross Taylor and Ruan Silva and Rui Hou and Rui Wang and Saghar Hosseini and Sa-hana Chennabasappa and Sanjay Singh and Sean Bell and Seohyun Sonia Kim and Sergey Edunov and Shaoliang Nie and Sharan Narang and Sharath Chandra Raparthy and Sheng Shen and Shengye Wan and Shruti Bhosale and Shun Zhang and Simon Vandenhende and Soumya Batra and Spencer Whit-man and Sten Sootla and St{\'e}phane Collot and Suchin Gururangan and Sydney Borodinsky and Tamar Herman and Tara Fowler and Tarek Sheasha and Thomas Georgiou and Thomas Scialom and Tobias Speckbacher and Todor Mihaylov and Tong Xiao and Ujjwal Karn and Vedanuj Goswami and Vibhor Gupta and Vignesh Ramanathan and Viktor Kerkez and Vincent Gonguet and Virginie Do and Vish Vogeti and Vladan Petrovic and Weiwei Chu and Wenhan Xiong and Wenyin Fu and Whit-ney Meers and Xavier Martinet and Xiaodong Wang and Xiaoqing Ellen Tan and Xinfeng Xie and Xuchao Jia and Xuewei Wang and Yaelle Goldschlag and Yashesh Gaur and Yasmine Babaei and Yiqian Wen and Yiwen Song and Yuchen Zhang and Yue Li and Yuning Mao and Zacharie Delpierre Coudert and Zhengxu Yan and Zhengxing Chen and Zoe Papakipos and Aaditya K. Singh and Aaron Grattafiori and Abha Jain and Adam Kelsey and Adam Shajnfeld and Adi Gangidi and Adolfo Victoria and Ahuva Goldstand and Ajay Menon and Ajay Sharma and Alex Boesenberg and Alex Vaughan and Alexei Baevski and Allie Fein-stein and Amanda Kallet and Amit Sangani and Anam Yunus and Andrei Lupu and Andres Alvarado and Andrew Caples and Andrew Gu and Andrew Ho and Andrew Poulton and Andrew Ryan and Ankit Ramchandani and Annie Franco and Aparajita Saraf and Arkabandhu Chowdhury and Ashley Gabriel and Ashwin R. Bharambe and Assaf Eisenman and Azadeh Yazdan and Beau James and Ben Maurer and Benjamin Leonhardi and Po-Yao (Bernie) Huang and Beth Loyd and Beto de Paola and Bhargavi Paranjape and Bing Liu and Bo Wu and Boyu Ni and Braden Hancock and Bram Wasti and Brandon Spence and Brani Stojkovic and Brian Gamido and Britt Montalvo and Carl Parker and Carly Burton and Catalina Mejia and Changhan Wang and Changkyu Kim and Chao Zhou and Chester Hu and Ching-Hsiang Chu and Chris Cai and Chris Tindal and Christoph Feichtenhofer and Damon Civin and Dana Beaty and Daniel Kreymer and Shang-Wen Li and Danny Wyatt and David Adkins and David Xu and Davide Testuggine and Delia David and Devi Parikh and Diana Liskovich and Didem Foss and Dingkang Wang and Duc Le and Dustin Holland and Edward Dowling and Eissa Jamil and Elaine Montgomery and Eleonora Presani and Emily Hahn and Emily Wood and Erik Brinkman and Esteban Arcaute and Evan Dunbar and Evan Smoth-ers and Fei Sun and Felix Kreuk and Feng Tian and Firat Ozgenel and Francesco Caggioni and Francisco (Paco) Guzm{\'a}n and Frank J. Kanayet and Frank Seide and Gabriela Medina Florez and Gabriella Schwarz and Gada Badeer and Georgia Swee and Gil Halpern and Govind Thattai and Grant Herman and Grigory Sizov and Guangyi Zhang and Guna Lakshminarayanan and Hamid Shojanazeri and Han Zou and Hannah Wang and Han Zha and Haroun Habeeb and Harrison Rudolph and Helen Suk and Henry Aspegren and Hunter Goldman and Igor Molybog and Igor Tufanov and Irina-Elena Veliche and Itai Gat and Jake Weissman and James Geboski and James Kohli and Japhet Asher and Jean-Baptiste Gaya and Jeff Marcus and Jeff Tang and Jennifer Chan and Jenny Zhen and Jeremy Reizenstein and Jeremy Teboul and Jessica Zhong and Jian Jin and Jingyi Yang and Joe Cummings and Jon Carvill and Jon Shepard and Jonathan McPhie and Jonathan Torres and Josh Ginsburg and Junjie Wang and Kaixing(Kai) Wu and U KamHou and Karan Saxena and Karthik Prasad and Kartikay Khandelwal and Katayoun Zand and Kathy Matosich and Kaushik Veeraraghavan and Kelly Michelena and Keqian Li and Kun Huang and Kunal Chawla and Kushal Lakhotia and Kyle Huang and Lailin Chen and Lakshya Garg and A Lavender and Leandro Silva and Lee Bell and Lei Zhang and Liangpeng Guo and Licheng Yu and Liron Moshkovich and Luca Wehrstedt and Madian Khabsa and Manav Avalani and Manish Bhatt and Maria Tsimpoukelli and Martynas Mankus and Matan Hasson and Matthias Lennie and Matthias Reso and Maxim Groshev and Maxim Naumov and Maya Lathi and Meghan Keneally and Michael L. Seltzer and Michal Valko and Michelle Re-strepo and Mihir Patel and Mik Vyatskov and Mikayel Samvelyan and Mike Clark and Mike Macey and Mike Wang and Miquel Jubert Hermoso and Mo Metanat and Mohammad Rastegari and Mun-ish Bansal and Nandhini Santhanam and Natascha Parks and Natasha White and Navy-ata Bawa and Nayan Singhal and Nick Egebo and Nicolas Usunier and Nikolay Pavlovich Laptev and Ning Dong and Ning Zhang and Norman Cheng and Oleg Chernoguz and Olivia Hart and Omkar Salpekar and Ozlem Kalinli and Parkin Kent and Parth Parekh and Paul Saab and Pavan Balaji and Pe-dro Rittner and Philip Bontrager and Pierre Roux and Piotr Doll{\'a}r and Polina Zvyagina and Prashant Ratanchandani and Pritish Yuvraj and Qian Liang and Rachad Alao and Rachel Rodriguez and Rafi Ayub and Raghotham Murthy and Raghu Nayani and Rahul Mitra and Raymond Li and Rebekkah Hogan and Robin Battey and Rocky Wang and Ro-han Maheswari and Russ Howes and Ruty Rinott and Sai Jayesh Bondu and Samyak Datta and Sara Chugh and Sara Hunt and Sargun Dhillon and S. Yu. Sidorov and Satadru Pan and Saurabh Verma and Seiji Yamamoto and Sharadh Ramaswamy and Shaun Lindsay and Sheng Feng and Shenghao Lin and Shengxin Zha and Shiva Shankar and Shuqiang Zhang and Sinong Wang and Sneha Agarwal and Soji Sajuyigbe and Soumith Chintala and Stephanie Max and Stephen Chen and Steve Kehoe and Steve Satterfield and Sudarshan Govindaprasad and Sumit Kumar Gupta and Sung-Bae Cho and Sunny Virk and Suraj Subramanian and Sy Choudhury and Sydney Goldman and Tal Remez and Tamar Glaser and Tamara Best and Thilo Kohler and Thomas Robinson and Tianhe Li and Tianjun Zhang and Tim Matthews and Timothy Chou and Tzook Shaked and Varun Vontimitta and Victoria O Ajayi and Victoria Montanez and Vijai Mohan and Vinay Kumar and Vishal Mangla and Vlad Ionescu and Vlad Andrei Poenaru and Vlad T. Mihailescu and Vladimir Ivanov and Wei Li and Wenchen Wang and Wenwen Jiang and Wes Bouaziz and Will Constable and Xia Tang and Xiaofang Wang and Xiaojian Wu and Xiaolan Wang and Xide Xia and Xilun Wu and Xinbo Gao and Yanjun Chen and Ye Hu and Ye Jia and Ye Qi and Yenda Li and Yilin Zhang and Ying Zhang and Yossi Adi and Youngjin Nam and Yu Wang and Yuchen Hao and Yundi Qian and Yuzi He and Zach Rait and Zachary DeVito and Zef Rosnbrick and Zhaoduo Wen and Zhenyu Yang and Zhiwei Zhao},
  year={2024},
  eprint={2407.21783},
  archivePrefix={arXiv},
  primaryClass={cs.AI},
  url={https://arxiv.org/abs/2407.21783}, 
}

@inproceedings{liu2026spatial,
  title={Spatial-ssrl: Enhancing spatial understanding via self-supervised reinforcement learning},
  author={Liu, Yuhong and Zhang, Beichen and Zang, Yuhang and Cao, Yuhang and Xing, Long and Dong, Xiaoyi and Duan, Haodong and Lin, Dahua and Wang, Jiaqi},
  booktitle={Proceedings of the IEEE/CVF Conference on Computer Vision and Pattern Recognition},
  pages={9570--9581},
  year={2026}
}
